\documentclass{ecai} 

\usepackage{latexsym}
\usepackage{amssymb}
\usepackage{amsmath}
\usepackage{amsthm}
\usepackage{booktabs}
\usepackage{enumitem}
\usepackage{graphicx}
\usepackage{color}
\usepackage{xspace}
\usepackage{pifont}
\usepackage{makecell}
\usepackage{algorithm}
\usepackage{algorithmic}
\usepackage{nicefrac}
\usepackage{mathtools}
\usepackage{tikz}
\usepackage{overpic}
\usepackage{listings}
\usepackage{xcolor}
\usepackage{silence}
\usepackage[font=footnotesize,format=plain,labelfont=bf,textfont=bf]{caption}
\usepackage[font=footnotesize]{subcaption}
\usepackage[bookmarksnumbered,unicode]{hyperref}
\hypersetup{hidelinks,pdflang={en},pdfdisplaydoctitle}
\usepackage{scalerel}
\usepackage{tikz}
\usetikzlibrary{svg.path}

\definecolor{cf09021}{RGB}{240,144,33}
\definecolor{cefb61d}{RGB}{239,182,29}
\definecolor{ccccd2b}{RGB}{204,205,43}
\definecolor{c07a6aa}{RGB}{7,166,170}

\tikzset{
  kaustlogo/.pic={
  \path[fill=cf09021] (2.8502, 1.6591).. controls (2.9163, 1.7583) and (2.9494, 1.8812) .. (2.9447, 2.0041).. controls (2.9447, 2.0797) and (2.9353, 2.1554) .. (2.9116, 2.231).. controls (2.836, 2.1648) and (2.7935, 2.0703) .. (2.7935, 1.971).. controls (2.7982, 1.8623) and (2.8171, 1.7583) .. (2.8502, 1.6591);

  \path[fill=cf09021] (3.6632, 2.4673).. controls (3.6868, 2.3444) and (3.6962, 2.2215) .. (3.6962, 2.0986).. controls (3.6962, 1.6401) and (3.6064, 1.314) .. (3.4694, 1.0777).. controls (3.493, 1.073) and (3.5166, 1.073) .. (3.5403, 1.073).. controls (4.1831, 1.073) and (4.3202, 1.6685) .. (4.3202, 1.9285).. controls (4.3202, 2.3491) and (3.8948, 2.4579) .. (3.6632, 2.4673);

  \path[fill=cf09021] (3.0156, 1.9994).. controls (3.0203, 1.8481) and (2.9731, 1.7016) .. (2.8785, 1.5834).. controls (2.9022, 1.522) and (2.9353, 1.4653) .. (2.9683, 1.4085).. controls (3.096, 1.5551) and (3.1574, 1.763) .. (3.1574, 2.0325).. controls (3.1574, 2.1412) and (3.1385, 2.2499) .. (3.1054, 2.3491).. controls (3.0581, 2.3255) and (3.0156, 2.3019) .. (2.9731, 2.2688).. controls (3.0014, 2.1884) and (3.0156, 2.0939) .. (3.0156, 1.9994);

  \path[fill=cf09021] (3.2378, 2.0325).. controls (3.2378, 1.7016) and (3.148, 1.4842) .. (3.0156, 1.3424).. controls (3.0487, 1.2998) and (3.0912, 1.262) .. (3.1338, 1.2242).. controls (3.2756, 1.4038) and (3.3701, 1.6685) .. (3.3701, 2.0655).. controls (3.3701, 2.1884) and (3.3559, 2.3066) .. (3.3228, 2.4248).. controls (3.2756, 2.4106) and (3.2283, 2.3964) .. (3.1858, 2.3822).. controls (3.2189, 2.2735) and (3.2425, 2.1554) .. (3.2378, 2.0325);

  \path[fill=cf09021] (3.4694, 2.0655).. controls (3.4694, 1.6401) and (3.3654, 1.3566) .. (3.2141, 1.1628).. controls (3.2567, 1.1344) and (3.3039, 1.1155) .. (3.3512, 1.1013).. controls (3.4883, 1.3235) and (3.5781, 1.6449) .. (3.5781, 2.0986).. controls (3.5781, 2.2215) and (3.5639, 2.3444) .. (3.5403, 2.4673).. controls (3.5025, 2.4626) and (3.4599, 2.4579) .. (3.4126, 2.4484).. controls (3.4505, 2.3255) and (3.4694, 2.1979) .. (3.4694, 2.0655);

  \path[fill=cefb61d] (2.1837, 3.0298).. controls (2.5382, 3.0298) and (2.8313, 2.8927) .. (3.0203, 2.6706).. controls (3.0581, 2.7226) and (3.0912, 2.7793) .. (3.1196, 2.836).. controls (2.9494, 3.0487) and (2.6895, 3.1905) .. (2.335, 3.1905).. controls (2.0466, 3.181) and (1.763, 3.129) .. (1.4889, 3.0298).. controls (1.4936, 2.992) and (1.5078, 2.9542) .. (1.5267, 2.9211).. controls (1.7394, 2.992) and (1.9616, 3.0298) .. (2.1837, 3.0298);

  \path[fill=cefb61d] (3.2141, 3.3039).. controls (3.1905, 3.6253) and (3.0203, 4.1547) .. (2.4295, 4.1547).. controls (1.971, 4.1547) and (1.6969, 3.7246) .. (1.5692, 3.3937).. controls (1.867, 3.4883) and (2.1743, 3.5403) .. (2.4815, 3.545).. controls (2.7462, 3.5544) and (3.0062, 3.4694) .. (3.2141, 3.3039);

  \path[fill=cefb61d] (1.6732, 2.7367).. controls (1.7914, 2.7651) and (1.9143, 2.7745) .. (2.0372, 2.7745).. controls (2.2357, 2.7745) and (2.6186, 2.7367) .. (2.8407, 2.4957).. controls (2.888, 2.5288) and (2.9305, 2.5618) .. (2.9683, 2.6044).. controls (2.7982, 2.8171) and (2.5193, 2.9494) .. (2.1837, 2.9494).. controls (1.9757, 2.9494) and (1.7678, 2.9163) .. (1.5645, 2.8502).. controls (1.5976, 2.8076) and (1.6354, 2.7698) .. (1.6732, 2.7367);

  \path[fill=cefb61d] (2.7793, 2.4626).. controls (2.5713, 2.6753) and (2.2215, 2.7084) .. (2.0419, 2.7084).. controls (1.9427, 2.7084) and (1.8434, 2.6989) .. (1.7489, 2.6847).. controls (1.9757, 2.5193) and (2.3019, 2.4011) .. (2.5146, 2.4011).. controls (2.5997, 2.3964) and (2.6942, 2.4153) .. (2.7793, 2.4626);

  \path[fill=cefb61d] (2.335, 3.2898).. controls (2.7036, 3.2898) and (2.9778, 3.148) .. (3.1621, 2.9353).. controls (3.1858, 3.0062) and (3.2047, 3.0771) .. (3.2141, 3.1527).. controls (3.0534, 3.3181) and (2.8171, 3.4316) .. (2.4815, 3.4316).. controls (2.1554, 3.4221) and (1.8292, 3.3607) .. (1.522, 3.2567).. controls (1.5125, 3.2189) and (1.5031, 3.1763) .. (1.4936, 3.1385).. controls (1.763, 3.233) and (2.0466, 3.285) .. (2.335, 3.2898);

  \path[fill=ccccd2b] (0.7894, 2.3964).. controls (0.7894, 2.6658) and (0.9595, 2.8596) .. (1.1958, 3.0014).. controls (1.1628, 3.0251) and (1.1249, 3.044) .. (1.0871, 3.0534).. controls (0.865, 2.9116) and (0.7137, 2.7178) .. (0.7137, 2.472).. controls (0.7137, 2.2499) and (0.7941, 1.9663) .. (0.9406, 1.6921).. controls (0.9642, 1.6921) and (0.9879, 1.6969) .. (1.0115, 1.7063).. controls (0.8697, 1.9474) and (0.7894, 2.2026) .. (0.7894, 2.3964);

  \path[fill=ccccd2b] (0.9784, 2.3208).. controls (0.9784, 2.5524) and (1.1297, 2.7178) .. (1.3329, 2.8313).. controls (1.3187, 2.8644) and (1.2951, 2.8974) .. (1.2715, 2.9258).. controls (1.0493, 2.8029) and (0.8886, 2.6327) .. (0.8886, 2.3917).. controls (0.8886, 2.2073) and (0.9642, 1.971) .. (1.0966, 1.7394).. controls (1.1202, 1.7583) and (1.1391, 1.7725) .. (1.158, 1.7961).. controls (1.0446, 1.9852) and (0.9831, 2.1743) .. (0.9784, 2.3208);

  \path[fill=ccccd2b] (0.9312, 3.0865).. controls (0.9028, 3.0912) and (0.8744, 3.0912) .. (0.8461, 3.0912).. controls (0.5767, 3.0912) and (0.0, 2.8691) .. (0.0, 2.4579).. controls (0.0, 2.0183) and (0.4868, 1.7489) .. (0.8035, 1.6969).. controls (0.6665, 1.9757) and (0.5908, 2.2546) .. (0.5908, 2.472).. controls (0.5956, 2.7273) and (0.7279, 2.9305) .. (0.9312, 3.0865);

  \path[fill=ccccd2b] (1.1628, 2.2499).. controls (1.1628, 2.4011) and (1.2431, 2.5335) .. (1.3849, 2.6233).. controls (1.3849, 2.6706) and (1.3755, 2.7178) .. (1.3613, 2.7604).. controls (1.1817, 2.6611) and (1.0588, 2.5193) .. (1.0588, 2.3255).. controls (1.0682, 2.1648) and (1.1202, 2.0088) .. (1.2053, 1.867).. controls (1.2242, 1.8954) and (1.2384, 1.9285) .. (1.2526, 1.9568).. controls (1.2006, 2.0466) and (1.1675, 2.1459) .. (1.1628, 2.2499);

  \path[fill=ccccd2b] (1.2337, 2.2499).. controls (1.2384, 2.1743) and (1.2573, 2.1034) .. (1.2904, 2.0372).. controls (1.3471, 2.1979) and (1.3802, 2.3681) .. (1.3896, 2.5382).. controls (1.2904, 2.472) and (1.2337, 2.3633) .. (1.2337, 2.2499);

  \path[fill=c07a6aa] (2.2593, 1.366).. controls (1.9663, 1.3707) and (1.6874, 1.4842) .. (1.4794, 1.6827).. controls (1.4416, 1.6685) and (1.4085, 1.6543) .. (1.3755, 1.6354).. controls (1.5976, 1.3755) and (1.9237, 1.1628) .. (2.3161, 1.1628).. controls (2.4815, 1.158) and (2.6422, 1.1911) .. (2.7887, 1.262).. controls (2.7651, 1.3282) and (2.732, 1.3896) .. (2.6895, 1.4416).. controls (2.5477, 1.3896) and (2.4059, 1.3613) .. (2.2593, 1.366);

  \path[fill=c07a6aa] (1.0021, 1.3755).. controls (0.865, 1.2384) and (0.7799, 1.0871) .. (0.7799, 0.9312).. controls (0.7799, 0.3356) and (1.7063, 0.0) .. (2.0655, 0.0).. controls (2.4909, 0.0) and (2.6895, 0.2694) .. (2.7745, 0.553).. controls (2.6564, 0.5247) and (2.5335, 0.5105) .. (2.4106, 0.5105).. controls (1.815, 0.5152) and (1.3235, 0.9075) .. (1.0021, 1.3755);

  \path[fill=c07a6aa] (1.919, 1.7772).. controls (1.7914, 1.7772) and (1.6685, 1.7536) .. (1.5503, 1.711).. controls (1.7441, 1.5362) and (1.9994, 1.4369) .. (2.2593, 1.4369).. controls (2.387, 1.4322) and (2.5146, 1.4558) .. (2.6327, 1.4983).. controls (2.4059, 1.73) and (2.0514, 1.7772) .. (1.919, 1.7772);

  \path[fill=c07a6aa] (2.3113, 1.0824).. controls (1.8907, 1.0824) and (1.5362, 1.3187) .. (1.2998, 1.5976).. controls (1.2715, 1.5834) and (1.2478, 1.5692) .. (1.2195, 1.5503).. controls (1.4794, 1.1958) and (1.8812, 0.8981) .. (2.3586, 0.8981).. controls (2.5193, 0.8933) and (2.68, 0.9264) .. (2.8265, 0.9879).. controls (2.8265, 1.054) and (2.8171, 1.1202) .. (2.8029, 1.1817).. controls (2.6469, 1.1155) and (2.4815, 1.0777) .. (2.3113, 1.0824);

  \path[fill=c07a6aa] (2.3586, 0.7988).. controls (1.8481, 0.7988) and (1.418, 1.1202) .. (1.1391, 1.4983).. controls (1.1202, 1.4842) and (1.1013, 1.4653) .. (1.0824, 1.4511).. controls (1.3849, 1.0068) and (1.8481, 0.6334) .. (2.4106, 0.6334).. controls (2.5429, 0.6334) and (2.6753, 0.6523) .. (2.8029, 0.6901).. controls (2.8171, 0.7563) and (2.8218, 0.8224) .. (2.8265, 0.8886).. controls (2.68, 0.8272) and (2.5193, 0.7988) .. (2.3586, 0.7988);
}
}

\newcommand\kausticon{\mbox{\scalerel*{
\begin{tikzpicture}[yscale=1,transform shape]
\pic{kaustlogo};
\end{tikzpicture}
}{|}}}

\definecolor{c0081fb}{RGB}{0,129,251}

\tikzset{
  mbzuailogo/.pic={
  \path[fill=c0081fb,nonzero rule,cm={ 1.3333,-0.0,-0.0,-1.3333,(3.4268, 8.0771)}] (0.0, 4.6885).. controls (-0.0735, 4.615) and (-0.1877, 4.6084) .. (-0.2684, 4.6677) -- (-0.3881, 4.5482).. controls (-0.3602, 4.5111) and (-0.3431, 4.4655) .. (-0.3431, 4.4156).. controls (-0.3431, 4.2933) and (-0.4422, 4.1941) .. (-0.5646, 4.1941).. controls (-0.6869, 4.1941) and (-0.7862, 4.2933) .. (-0.7862, 4.4156).. controls (-0.7862, 4.5379) and (-0.6869, 4.6371) .. (-0.5646, 4.6371).. controls (-0.5147, 4.6371) and (-0.4691, 4.6199) .. (-0.4321, 4.5921) -- (-0.3124, 4.7118).. controls (-0.3647, 4.783) and (-0.3656, 4.8797) .. (-0.3148, 4.9521) -- (-0.4343, 5.0657).. controls (-0.4699, 5.0398) and (-0.5149, 5.0236) .. (-0.5646, 5.0236).. controls (-0.6786, 5.0236) and (-0.771, 5.1053) .. (-0.771, 5.2061).. controls (-0.771, 5.3071) and (-0.6786, 5.3889) .. (-0.5646, 5.3889).. controls (-0.4507, 5.3889) and (-0.3583, 5.3071) .. (-0.3583, 5.2061).. controls (-0.3583, 5.1706) and (-0.3703, 5.1377) .. (-0.3901, 5.1096) -- (-0.2715, 4.9968).. controls (-0.1901, 5.0601) and (-0.0733, 5.0556) .. (0.001, 4.9813).. controls (0.0816, 4.9007) and (0.0813, 4.7695) .. (0.0, 4.6885);

  \path[fill=c0081fb,nonzero rule,cm={ 1.3333,-0.0,-0.0,-1.3333,(2.346, 9.1503)}] (0.0, 4.6885) -- (0.1268, 4.5747).. controls (0.1606, 4.596) and (0.2015, 4.6087) .. (0.246, 4.6087).. controls (0.2906, 4.6087) and (0.3316, 4.596) .. (0.3653, 4.5747) -- (0.4925, 4.6887).. controls (0.4418, 4.7674) and (0.4517, 4.8744) .. (0.5219, 4.9446).. controls (0.5919, 5.0145) and (0.6983, 5.0245) .. (0.7768, 4.9746) -- (0.8885, 5.1004).. controls (0.867, 5.1343) and (0.8539, 5.1756) .. (0.8539, 5.2204).. controls (0.8539, 5.3344) and (0.9358, 5.4268) .. (1.0367, 5.4268).. controls (1.1374, 5.4268) and (1.2192, 5.3344) .. (1.2192, 5.2204).. controls (1.2192, 5.1064) and (1.1374, 5.0141) .. (1.0367, 5.0141).. controls (0.9967, 5.0141) and (0.96, 5.029) .. (0.9299, 5.0535) -- (0.8227, 4.9328).. controls (0.8883, 4.8532) and (0.8837, 4.7347) .. (0.8077, 4.6587).. controls (0.7322, 4.5832) and (0.6142, 4.5782) .. (0.5346, 4.6431) -- (0.4123, 4.5335).. controls (0.4373, 4.5033) and (0.4524, 4.4663) .. (0.4524, 4.4261).. controls (0.4524, 4.3252) and (0.36, 4.2435) .. (0.246, 4.2435).. controls (0.1321, 4.2435) and (0.0397, 4.3252) .. (0.0397, 4.4261).. controls (0.0397, 4.4663) and (0.0548, 4.5033) .. (0.0798, 4.5335) -- (-0.0424, 4.643).. controls (-0.1152, 4.5829) and (-0.2177, 4.5787) .. (-0.2931, 4.6314) -- (-0.3719, 4.5493).. controls (-0.3269, 4.4904) and (-0.3185, 4.4158) .. (-0.3476, 4.3578) -- (-0.2433, 4.2694).. controls (-0.1802, 4.3184) and (-0.083, 4.3131) .. (-0.0128, 4.2522).. controls (0.0659, 4.1838) and (0.0806, 4.0719) .. (0.0202, 4.0023).. controls (-0.0104, 3.967) and (-0.0544, 3.9493) .. (-0.1006, 3.9493).. controls (-0.1457, 3.9493) and (-0.1929, 3.9662) .. (-0.2317, 4.0).. controls (-0.3008, 4.0598) and (-0.3205, 4.1531) .. (-0.2837, 4.2221) -- (-0.3867, 4.3093).. controls (-0.4537, 4.2549) and (-0.5585, 4.2653) .. (-0.6274, 4.3364).. controls (-0.7001, 4.4112) and (-0.7052, 4.5238) .. (-0.6391, 4.5882).. controls (-0.5808, 4.6446) and (-0.4871, 4.6448) .. (-0.4159, 4.5933) -- (-0.3376, 4.6748).. controls (-0.3997, 4.7559) and (-0.3937, 4.8725) .. (-0.3187, 4.9476).. controls (-0.2371, 5.0292) and (-0.1056, 5.0299) .. (-0.025, 4.9495).. controls (0.0455, 4.879) and (0.0537, 4.7695) .. (0.0, 4.6885);

  \path[fill=c0081fb,nonzero rule,cm={ 1.3333,-0.0,-0.0,-1.3333,(1.6149, 8.7162)}] (0.0, 4.6885).. controls (-0.1009, 4.6885) and (-0.1827, 4.7808) .. (-0.1827, 4.8949).. controls (-0.1827, 4.9413) and (-0.1686, 4.9838) .. (-0.1456, 5.0183) -- (-0.2292, 5.1167).. controls (-0.2593, 5.0939) and (-0.2956, 5.081) .. (-0.3343, 5.0838).. controls (-0.3634, 5.0861) and (-0.3896, 5.0972) .. (-0.4123, 5.1138) -- (-0.5157, 4.9906).. controls (-0.5022, 4.9626) and (-0.4941, 4.9301) .. (-0.4941, 4.8949).. controls (-0.4941, 4.8589) and (-0.5024, 4.8255) .. (-0.5165, 4.7973) -- (-0.4175, 4.6761).. controls (-0.3952, 4.692) and (-0.3696, 4.7026) .. (-0.3413, 4.7048).. controls (-0.2494, 4.7117) and (-0.1685, 4.633) .. (-0.1607, 4.529).. controls (-0.1528, 4.4251) and (-0.2212, 4.3353) .. (-0.3132, 4.3285).. controls (-0.3411, 4.3263) and (-0.3677, 4.3329) .. (-0.3918, 4.345) -- (-0.4501, 4.2454).. controls (-0.445, 4.239) and (-0.4399, 4.2325) .. (-0.4353, 4.2254).. controls (-0.3875, 4.1514) and (-0.3888, 4.0704) .. (-0.4384, 4.0447).. controls (-0.4487, 4.0394) and (-0.4603, 4.0368) .. (-0.4726, 4.0368).. controls (-0.5194, 4.0368) and (-0.5769, 4.0736) .. (-0.6148, 4.1324).. controls (-0.6627, 4.2066) and (-0.6614, 4.2874) .. (-0.6119, 4.313).. controls (-0.58, 4.3296) and (-0.5362, 4.3191) .. (-0.496, 4.29) -- (-0.4417, 4.3826).. controls (-0.4705, 4.414) and (-0.4901, 4.4561) .. (-0.4938, 4.5042).. controls (-0.4973, 4.5517) and (-0.4843, 4.596) .. (-0.461, 4.631) -- (-0.5547, 4.7456).. controls (-0.5779, 4.7247) and (-0.606, 4.7121) .. (-0.6366, 4.7121).. controls (-0.7152, 4.7121) and (-0.7789, 4.794) .. (-0.7789, 4.8949).. controls (-0.7789, 4.9959) and (-0.7152, 5.0777) .. (-0.6366, 5.0777).. controls (-0.6053, 5.0777) and (-0.5767, 5.0644) .. (-0.5532, 5.0426) -- (-0.4552, 5.1594).. controls (-0.4778, 5.1942) and (-0.4903, 5.2377) .. (-0.4868, 5.2846).. controls (-0.4789, 5.3884) and (-0.398, 5.467) .. (-0.3061, 5.4601).. controls (-0.2142, 5.4532) and (-0.146, 5.3635) .. (-0.1537, 5.2595).. controls (-0.1564, 5.2231) and (-0.1685, 5.1901) .. (-0.1865, 5.1626) -- (-0.1035, 5.0648).. controls (-0.074, 5.0876) and (-0.0384, 5.1012) .. (-0.0, 5.1012).. controls (0.1009, 5.1012) and (0.1827, 5.0088) .. (0.1827, 4.8949).. controls (0.1827, 4.7808) and (0.1009, 4.6885) .. (0.0, 4.6885);

  \path[fill=c0081fb,nonzero rule,cm={ 1.3333,-0.0,-0.0,-1.3333,(4.5884, 8.6847)}] (0.0, 4.6885).. controls (-0.0313, 4.6885) and (-0.0597, 4.7028) .. (-0.0828, 4.7261) -- (-0.1855, 4.6055).. controls (-0.1635, 4.571) and (-0.1512, 4.528) .. (-0.1547, 4.4816).. controls (-0.1581, 4.4368) and (-0.1756, 4.3973) .. (-0.2012, 4.3667) -- (-0.1524, 4.2917).. controls (-0.1103, 4.3261) and (-0.0626, 4.3381) .. (-0.0291, 4.317).. controls (0.0191, 4.2865) and (0.0193, 4.2) .. (-0.0288, 4.1238).. controls (-0.0654, 4.0656) and (-0.1194, 4.0302) .. (-0.1645, 4.0302).. controls (-0.1785, 4.0302) and (-0.1916, 4.0336) .. (-0.203, 4.0408).. controls (-0.2514, 4.0712) and (-0.2515, 4.1577) .. (-0.2035, 4.234).. controls (-0.2011, 4.2378) and (-0.1984, 4.2409) .. (-0.1959, 4.2445) -- (-0.2492, 4.3264).. controls (-0.2751, 4.3117) and (-0.3045, 4.3038) .. (-0.3353, 4.3061).. controls (-0.4274, 4.3129) and (-0.4955, 4.4027) .. (-0.4877, 4.5067).. controls (-0.4799, 4.6106) and (-0.3991, 4.6892) .. (-0.3071, 4.6823).. controls (-0.2775, 4.68) and (-0.2509, 4.6687) .. (-0.228, 4.6516) -- (-0.1185, 4.7801).. controls (-0.1303, 4.807) and (-0.1375, 4.8379) .. (-0.1375, 4.8712).. controls (-0.1375, 4.9723) and (-0.0759, 5.0541) .. (0.0, 5.0541).. controls (0.076, 5.0541) and (0.1376, 4.9723) .. (0.1376, 4.8712).. controls (0.1376, 4.7703) and (0.076, 4.6885) .. (0.0, 4.6885);

  \path[fill=c0081fb,nonzero rule,cm={ 1.3333,-0.0,-0.0,-1.3333,(1.4138, 9.7618)}] (0.0, 4.6885).. controls (0.0465, 4.6459) and (0.0707, 4.592) .. (0.0681, 4.5484) -- (0.1858, 4.4637).. controls (0.2184, 4.4758) and (0.2644, 4.4727) .. (0.3115, 4.4511).. controls (0.3898, 4.4152) and (0.4408, 4.343) .. (0.4255, 4.2899).. controls (0.4159, 4.2565) and (0.3826, 4.2385) .. (0.3399, 4.2385).. controls (0.3143, 4.2385) and (0.2854, 4.245) .. (0.2561, 4.2584).. controls (0.1783, 4.294) and (0.1274, 4.3655) .. (0.1418, 4.4186) -- (0.0402, 4.4919).. controls (-0.0018, 4.4594) and (-0.0792, 4.475) .. (-0.1407, 4.5315).. controls (-0.2072, 4.5923) and (-0.2295, 4.6768) .. (-0.1906, 4.7201).. controls (-0.1517, 4.7635) and (-0.0664, 4.7493) .. (0.0, 4.6885);

  \path[fill=c0081fb,nonzero rule,cm={ 1.3333,-0.0,-0.0,-1.3333,(4.5117, 7.6722)}] (0.0, 4.6885).. controls (-0.0301, 4.6744) and (-0.0701, 4.6843) .. (-0.1071, 4.7105) -- (-0.1484, 4.6312).. controls (-0.1162, 4.5995) and (-0.0943, 4.5542) .. (-0.0905, 4.5027).. controls (-0.0827, 4.3988) and (-0.1509, 4.3089) .. (-0.2428, 4.302).. controls (-0.2466, 4.3018) and (-0.2503, 4.3016) .. (-0.254, 4.3016).. controls (-0.3413, 4.3016) and (-0.416, 4.378) .. (-0.4235, 4.4777).. controls (-0.4266, 4.5203) and (-0.4165, 4.5601) .. (-0.3978, 4.5932) -- (-0.5195, 4.7211).. controls (-0.5864, 4.6804) and (-0.6808, 4.6949) .. (-0.7444, 4.7606).. controls (-0.817, 4.8354) and (-0.8222, 4.9481) .. (-0.756, 5.0123).. controls (-0.7014, 5.0653) and (-0.6155, 5.0679) .. (-0.5464, 5.0255) -- (-0.4564, 5.1389).. controls (-0.5074, 5.1971) and (-0.5262, 5.2648) .. (-0.4974, 5.2982).. controls (-0.4864, 5.3109) and (-0.4701, 5.317) .. (-0.4508, 5.3171) -- (-0.4493, 5.3171).. controls (-0.4118, 5.3168) and (-0.3632, 5.2944) .. (-0.3204, 5.2541).. controls (-0.255, 5.1924) and (-0.2281, 5.1121) .. (-0.2603, 5.0748).. controls (-0.2883, 5.0424) and (-0.3515, 5.0531) .. (-0.4105, 5.0966) -- (-0.4986, 4.9855).. controls (-0.4399, 4.9193) and (-0.431, 4.8276) .. (-0.4746, 4.7641) -- (-0.3582, 4.6418).. controls (-0.3337, 4.6624) and (-0.3041, 4.6758) .. (-0.2711, 4.6783).. controls (-0.2462, 4.6802) and (-0.2224, 4.6754) .. (-0.2004, 4.666) -- (-0.154, 4.7551).. controls (-0.1609, 4.7636) and (-0.1673, 4.7728) .. (-0.1733, 4.7826).. controls (-0.2187, 4.858) and (-0.215, 4.9381) .. (-0.1649, 4.9617).. controls (-0.1148, 4.9852) and (-0.0373, 4.9431) .. (0.0083, 4.8676).. controls (0.0538, 4.7923) and (0.0501, 4.712) .. (0.0, 4.6885);
}
}

\newcommand\mbzuaiicon{\mbox{\scalerel*{
\begin{tikzpicture}[yscale=1,transform shape]
\pic{mbzuailogo};
\end{tikzpicture}
}{|}}}

\newcommand{\iconbox}[1]{\raisebox{0pt}[0pt][0pt]{\textsuperscript{\footnotesize #1}}}

\newcommand{\powersgd}{PowerSGD\xspace}
\newcommand{\GradiVeQ}{GradiVeQ\xspace}
\newcommand{\intsgd}{IntSGD\xspace}
\newcommand{\THC}{THC\xspace}
\newcommand{\signsgd}{SignSGD\xspace}
\newcommand{\cmark}{{\normalsize {\color{green}\ding{51}}}}
\newcommand{\xmark}{{\normalsize {\color{red}\ding{55}}}}

\newcommand{\R}{\mathbb{R}}
\newcommand{\N}{\mathbb{N}}
\DeclarePairedDelimiterX{\norm}[1]{\lVert}{\rVert}{#1}
\DeclarePairedDelimiterX{\abs}[1]{\lvert}{\rvert}{#1}
\DeclarePairedDelimiterX{\sbr}[1]{[}{]}{#1}
\DeclarePairedDelimiterX{\cbr}[1]{\{}{\}}{#1}
\DeclarePairedDelimiterX{\rbr}[1]{(}{)}{#1}
\newcommand{\floor}[1]{\lfloor #1 \rfloor}
\newcommand{\signum}{\operatorname{sign}}
\newcommand{\sign}{\operatorname{sign}}

\providecommand{\circ}{\,\cdot\,}
\newcommand{\E}[1]{\mathbb{E}\left[#1\right]}
\newcommand{\cC}{\mathcal{C}}
\newcommand{\cG}{\mathcal{G}}
\newcommand{\C}{\mathcal{C}}
\newcommand{\cO}{\mathcal{O}}
\newcommand{\bx}{\mathbf{x}}

\newcommand{\Prob}{\operatorname{Pr}}
\newcommand{\EE}[2]{\mathbb{E}_{#1}\left[#2\right]}
\newcommand{\dotprod}[2]{\langle #1, #2 \rangle}
\newcommand{\cD}{\mathcal{D}}

\newcommand{\GQ}{\mathcal{Q}}
\newcommand{\GL}{\mathcal{L}}
\newcommand{\GE}{\mathcal{E}}
\newcommand{\cQ}{\mathcal{Q}}
\newcommand{\cL}{\mathcal{L}}
\newcommand{\cE}{\mathcal{E}}
\newcommand{\U}{\mathcal{U}}
\newcommand{\NC}{\mathcal{C}_{\text{nat}}}
\newcommand{\xx}{\mathbf{x}}
\newcommand{\yy}{\mathbf{y}}
\newcommand{\eqdef}{\mathrel{\mathop:}=}
\newcommand{\smartparagraph}[1]{\noindent\textbf{#1}}

\newcommand{\eonenotzero}{\operatorname{gz}_{1}}
\newcommand{\etwonotzero}{\operatorname{gz}_{2}}

\newcommand{\nonz}{\operatorname{non\_zero}}
\newcommand{\diff}{\operatorname{diff}}
\newcommand{\smaller}{\operatorname{smaller}}

\usepackage{url}

\newtheorem{theorem}{Theorem}
\newtheorem{lemma}[theorem]{Lemma}

\newtheorem{definition}{Definition}

\newcommand{\BibTeX}{B\kern-.05em{\sc i\kern-.025em b}\kern-.08em\TeX}

\newif\ifarxiv
\arxivtrue

\begin{document}


\begin{frontmatter}


\paperid{5578} 


\title{Quantize Once, Train Fast: Allreduce-Compatible Compression with Provable Guarantees}


\author[\iconbox{\kausticon}]{\fnms{Jihao}~\snm{Xin}}
\author[\iconbox{\kausticon}]{\fnms{Marco}~\snm{Canini}}
\author[\iconbox{\kausticon}]{\fnms{Peter}~\snm{Richtárik}}
\author[\iconbox{\mbzuaiicon}]{\fnms{Samuel}~\snm{Horváth}\thanks{Corresponding Author. Email: samuel.horvath@mbzuai.ac.ae}}

\address[\kausticon]{~KAUST}
\address[\mbzuaiicon]{~MBZUAI}


\begin{abstract}
    Distributed training enables large-scale deep learning, but suffers from high communication overhead, especially as models and datasets grow. Gradient compression, particularly quantization, is a promising approach to mitigate this bottleneck. However, existing quantization schemes are often incompatible with Allreduce, the dominant communication primitive in distributed deep learning, and many prior solutions rely on heuristics without theoretical guarantees.
    We introduce Global-QSGD, an Allreduce-compatible gradient quantization method that leverages global norm scaling to reduce communication overhead while preserving accuracy.
    Global-QSGD is backed by rigorous theoretical analysis, extending standard unbiased compressor frameworks to establish formal convergence guarantees. Additionally, we develop a performance model to evaluate its impact across different hardware configurations.
    Extensive experiments on NVLink, PCIe, and large-scale cloud environments show that Global-QSGD accelerates distributed training by up to 3.51× over baseline quantization methods, making it a practical and efficient solution for large-scale deep learning workloads.
\end{abstract}

\end{frontmatter}

\section{Introduction}
Distributed deep learning has become the standard approach for scaling training across multiple compute nodes, enabling faster convergence on large models and datasets~\citep{dean2012large}. However, as training scales up, communication overhead increasingly dominates total runtime, particularly in large-scale deployments. For example, \citet{switchML} reports that communication accounts for more than 90\% of the total training time in deep learning workloads, significantly limiting the benefits of additional computing resources.

Gradient quantization has emerged as a practical solution to alleviate this bottleneck, which reduces the gradient bit-width (e.g., from 32-bit to 8-bit), making it communication-efficient.
However, existing quantization methods such as QSGD~\citep{qsgd2017neurips} are often impractical in real-world scenarios due to their incompatibility with \textit{Allreduce}--the dominant communication primitive for distributed deep learning~\citep{agarwal2022utility,castello2023analyzing,langer2020distributed,narayanan2021efficient,xin2024immediate}.
The key issue arises because quantization is performed locally on each worker. For example, when reducing the precision from 32-bit to 8-bit, the gradients must be scaled using a \textit{norm}, which varies between workers. As a result, gradients are quantized at different scales, and directly aggregating these quantized values via Allreduce leads to numerical inconsistencies.

To address this limitation, we introduce \textbf{Global-QSGD}. Instead of using local norms for scaling, Global-QSGD leverages the \textit{global norm} computed across all workers, ensuring consistent quantization scales and enabling Allreduce compatibility; that is, quantized gradient values can be directly aggregated in their compressed (quantized) representation without the need (and overhead) of switching numeric representation.

Like other lossy compression methods, quantization inevitably loses information. Traditional quantization applies a linear scaling approach, known as \textbf{linear dithering}, which is suboptimal because gradients decrease in magnitude as the model converges. Smaller gradients require finer precision to maintain accuracy. To address this, we leverage \textbf{exponential dithering}, which has higher precision for smaller gradient values, improving convergence accuracy. 

Unlike previous approaches that rely on heuristics, we provide a rigorous convergence analysis (\S \ref{sec:variance}, \S \ref{sec:convergence}) to establish the theoretical foundations. Using its unbiased nature, we prove that Global-QSGD maintains a bounded variance, ensuring stable convergence and making it a theoretically sound choice.
To study the impact across different hardware configurations, we derive a performance model (\S \ref{sec:Communication Modeling}).

\noindent This paper makes the following key contributions:
\begin{itemize}
\item We introduce Global-QSGD, the first quantization method that seamlessly integrates with Allreduce.
\item We develop exponential dithering with a custom reduction function to effectively overcome the convergence limitations of traditional linear dithering.
\item We provide a comprehensive theoretical framework that rigorously establishes Global-QSGD's convergence guarantees.
\item We evaluate Global-QSGD across several settings, including single-node and cloud environments. Our results show that Global-QSGD accelerates model training by up to 3.51×, significantly reducing communication overhead without sacrificing model accuracy.
We openly release our code at \url{https://github.com/sands-lab/global-qsgd}.

\end{itemize}

\section{Related Work}
\begin{table}[tb]
\setlength\tabcolsep{3.5pt} 
    \centering
	\caption{
	 Comparison of Allreduce-compatible compressors.
	}
        \label{tab:comparasion}
	\centering 
        \scalebox{0.8}{
	\begin{tabular}{lccccc}\toprule[.1em]
		Algorithm & \makecell{Seamless \\ Integration} & \makecell{w/o\\Extra Step} & \makecell{Rigorous \\ Guarantees} & \makecell{w/o\\ EF}  & \makecell{Tunable \\ Compression} \\
		\midrule
            \powersgd~\citep{vogels2019powersgd}          & \xmark & \xmark & \xmark & \xmark & \cmark \\
            \GradiVeQ~\citep{GradiVeQ}          & \xmark & \xmark & \xmark & \cmark & \cmark \\
		\intsgd~\citep{mishchenko2021intsgd}            & \cmark & \cmark & \xmark & \cmark & \xmark \\
            \THC~\citep{THC}             & \xmark & \xmark & \xmark & \xmark & \cmark \\
            Global-QSGD        & \cmark & \cmark & \cmark & \cmark & \cmark \\
		\bottomrule[.1em]
        \end{tabular}
        }
        %
\end{table} 
Gradient compression methods can be broadly categorized into three approaches: \textit{sparsification}~\citep{agarwal2022utility, alistarh2018sparse, RDME, OKTopK, stich2018sparsified}, \textit{decomposition}~\citep{vogels2019powersgd}, and \textit{quantization}~\citep{qsgd2017neurips, horvath2019natural, Na2017:limitedprecision, 1bit, terngrad}.
Although these methods differ in how they reduce communication overhead, they can also be classified according to their impact on gradient updates, falling into two major theoretical categories: \textit{unbiased} and \textit{biased} compressors. Unbiased compressors, such as quantization schemes like QSGD~\citep{qsgd2017neurips}, maintain an expectation-equal gradient estimate with bounded variance, making them easier to analyze in theoretical frameworks~\citep{gorbunov2020unified}. In contrast, biased compressors, including Top-$k$~\citep{alistarh2018sparse}, \signsgd~\citep{bernstein2018signsgd}, and \powersgd~\citep{vogels2019powersgd}, introduce systematic distortion, requiring correction mechanisms such as error feedback (EF)~\citep{karimireddy2019error, EF21, stich2018sparsified} or induced compressors (IC)~\citep{horvath2021a} to restore convergence guarantees. However, these mechanisms introduce additional memory overhead or rely on an auxiliary unbiased compressor, which limits practicality.

Unfortunately, most of the existing compressors are not natively compatible with Allreduce since the sum of two compressed vectors is not a homomorphic operation; thus, it generally requires an expensive decompress-aggregate-compress operation. This includes greedy and random sparsification,\footnote{For random sparsification, one can share random seeds across workers to make it Allreduce compatible, e.g., see synchronized random seed~\citep{xie2020cser}.} quantization, and sign-based methods.

Several compressors have been proposed to ensure \textit{Allreduce} compatibility, but all rely on some heuristics, so they all lack rigorous theoretical guarantees.
\GradiVeQ~\citep{GradiVeQ} assumes adjacent gradients are linearly correlated, \intsgd~\citep{mishchenko2021intsgd} requires clipping the communicated integer values, and \powersgd approximates low-rank decomposition using power iteration while depending on memory-intensive error feedback.
\THC~\citep{THC}, developed in parallel to our work, also utilizes global norm. The Uniform THC variant is similar to linear dithering in our work. However, THC focuses on system efficiency to heuristically select the quantization scheme.
We focus on convergence properties with exponential dithering and provide theoretical guarantees.
Table~\ref{tab:comparasion} compares our work with the above methods and summarizes the key differences.




\begin{figure}[t!]
    \centering
    \includegraphics[width=\linewidth, height=0.3\linewidth]{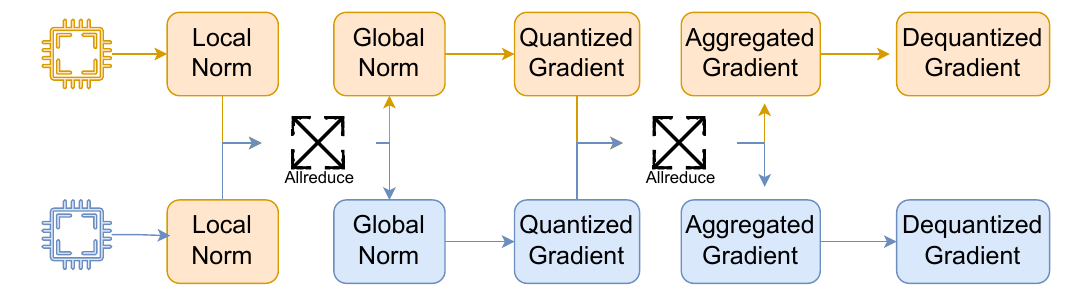} 
    \caption{Global-QSGD workflow.}
    \label{fig:workflow}
    \vspace{1.5em}
\end{figure}

\begin{figure}[t!]
    \centering
    \includegraphics[width=\linewidth]{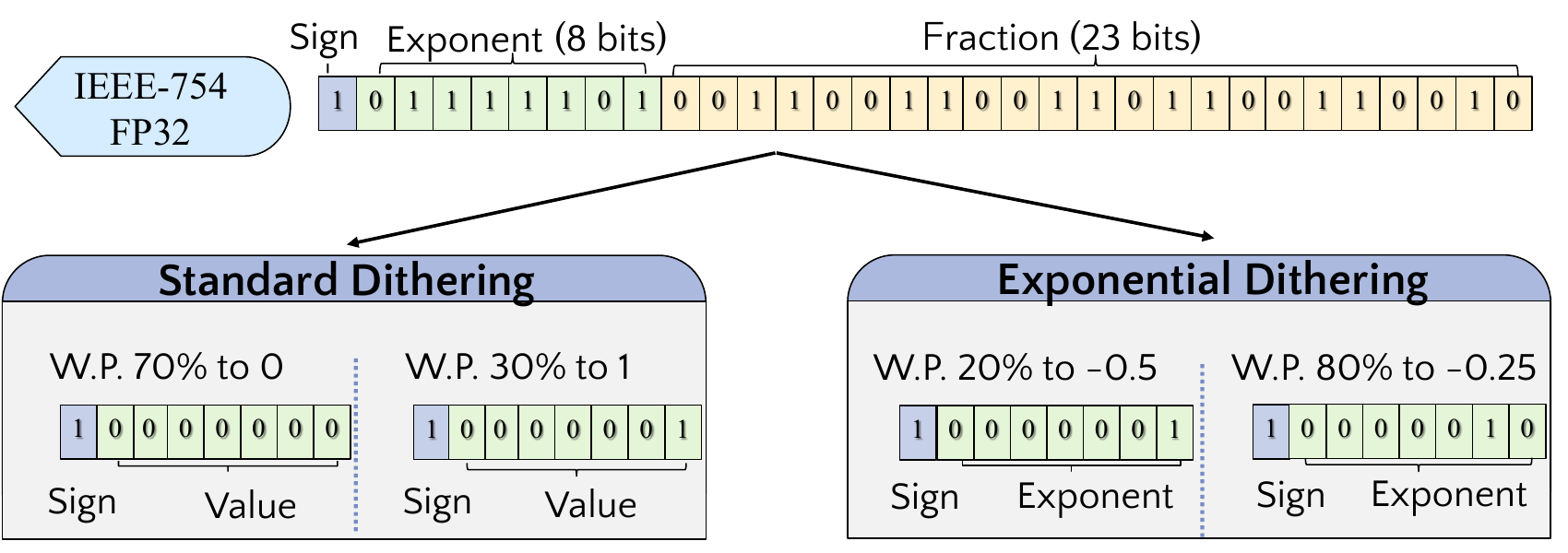}
    \caption{Quantize $-0.3$ from FP32 to 8-bit.}
    \label{fig:dithering}
    \vspace{1.5em}
\end{figure}

\section{Global-QSGD Overview}

Global-QSGD is a quantization operator~\citep{qsgd2017neurips, horvath2019natural}, where the main innovation is that we propose to normalize by the global norm across workers instead of each worker using its local norm.
Figure~\ref{fig:workflow} illustrates an example workflow of the Global-QSGD's two-round Allreduce with 2 workers.
Algorithm~\ref{alg:global_quantization} formalizes the high-level process of Global-QSGD, which can be categorized into 3 main steps:
\begin{algorithm}[th]
    \caption{Global-QSGD}
    \label{alg:global_quantization}
    \begin{algorithmic}[1]
    \REQUIRE Worker count: $n$, Gradients: $\xx = \sbr*{x_1, x_2, \hdots, x_n}$,\\
    $\textit{sparse} \in \cbr*{\text{True, False}}$
    \STATE $\norm{\xx}_{q,p} = \text{Allreduce}\rbr*{\text{MAX}\cbr*{\norm{x_i}_{q,p}}}$
    \FOR{$i \in \{1,2, \hdots, n\}$} 
        \STATE $y_i \eqdef |x_i|/\|\xx\|_{q,p}$ 
        \STATE $z_i \eqdef \signum(x_i) \circ \xi_i(y_i)$
    \ENDFOR
    \IF{\textit{sparse}}
        \RETURN $\frac{1}{n}\text{Allgather}(\text{SUM}\cbr{z_i})$
    \ENDIF
    \RETURN $\frac{1}{n}\text{Allreduce}(\text{SUM}\cbr{z_i})$
    \end{algorithmic}
\end{algorithm}

\smartparagraph{Step 1: Obtain global norm (line 1)}, where each worker computes its local norm and then performs an Allreduce operation (max as the reduction operator) to obtain the global norm. This ensures that all workers quantize their gradients consistently, preventing local scaling discrepancies.
As a one-element Allreduce, the overhead of this step is negligible, and can be hidden by overlapping this step with the gradient communication of a preceding communication round in a sequence of Allreduce invocations.

\smartparagraph{Step 2: Quantization (line 2-5)}, where each worker normalizes every gradient element to \([0,1]\) while preserving the sign bit, and maps each element to the nearest quantization level \( l_{i} \) for \( i \in \{0,1, \hdots, s\} \) with stochastic rounding.
Figure~\ref{fig:dithering} illustrates an example of how the value \( -0.3 \) represented in IEEE FP32 format is converted to an 8-bit representation. \S \ref{sec:quantization} details the quantization schemes.

\smartparagraph{Step 3: Aggregation (line 9)}, where the workers perform an Allreduce operation, applying the reduction operator directly on the quantized values without requiring changes of numerical representation.

\smartparagraph{Optional: Sparsity handling (lines 6-8)}. Additionally, we handle sparse gradients differently by only transmitting non-zero elements. Instead of sending full quantized vectors, we use Allgather to efficiently communicate the non-zero values.

\section{Quantization Schemes}
\label{sec:quantization}
\subsection{Formulation}
\label{sec:formulation}
To establish a solid theoretical foundation, we first present the mathematical formulation of the Global-QSGD quantization operator, which also serves as the foundation for our subsequent variance analysis (\S~\ref{sec:variance}) and convergence guarantees (\S~\ref{sec:convergence}).

We begin with the notations. For a vector $x_i \in \R^d$, we define the $\ell_p$-norm as $\norm{x_i}_p \eqdef (\sum_{j=1}^d|x_{ij}|^p)^{\nicefrac{1}{p}}$ for $p\in(1,\infty)$. When $p = \infty$, $\norm{x_i}_{\infty}$ represents the maximum absolute value among all elements of $x_i$. Let $\xx = \sbr*{x_1, x_2, \hdots, x_n} \in \R^{nd}$ be a concatenation of vectors $x_1, x_2, \hdots, x_n \in \R^d$. We define the $(q,p)$-mixed norm in $\R^{nd}$ as $\norm{\xx}_{q,p} \eqdef (\sum_{i=1}^n \norm{x_i}_q^p)^{\nicefrac{1}{p}}$.
For any vectors $x_i, x_j$, $x_i \circ x_j$ represents their element-wise multiplication, and, for any vector $x_i$, $\abs{x_i}, \sign(x_i)$ stand for element-wise absolute value and signum operations, respectively. We denote $[n] \eqdef \cbr*{1, 2, \hdots, n}$ for any $n \in \N$.
We summarize the notation in Table~\ref{tb:notation}.

\begin{table}[th]
    \begin{center}
    \caption{Summary of key notation.}
    \label{tb:notation}
    \vspace{6pt}
    \begin{tabular}{|l|l|}
    \hline 
    \textbf{Symbol} & \textbf{Description} \\ \hline
    $d$ & Dimension of gradients\\
    $n$ & Number of workers \\
    $s$ & Number of quantization levels\\
    $l_i$ & Value of the $i_{th}$ level\\
    $x_i  \in \R^{d}$ & Gradients of the $i_{th}$ worker\\
    $y_i  \in \R^{d}$ & Normalized gradients of the $i_{th}$ worker\\
    $\xx \in \R^{nd}$ & Gradients of all $n$ workers\\
    $\yy \in \R^{nd}$ & Normalized gradients of all $n$ workers\\
    $\norm{x_i}_p$ & $\ell_p$ norm of $x_i$\\
    $\norm{\xx}_{q,p}$ & $(q,p)$-mixed norm of $\xx$\\
    $\xi_i(y_i)$ & Random rounding of $y_i$\\
    $\GQ^{q,p}_s$ & Global-QSGD compressor in general\\
    $\GL^{q,p}_s$ & Global-QSGD with linear dithering\\
    $\GE^{q,p}_s$ & Global-QSGD with exponential dithering\\\hline
    \end{tabular}
    \end{center}
    \end{table}

We define the Global-QSGD quantization operator:
 \begin{definition}[$\GQ^{q,p}_s$] The Global-QSGD quantization operator with respect to the $(q,p)$-mixed norm and with $s$ levels 
 $$0 = l_s <l_{s-1} < l_{s-2} < \dots < l_{1} < l_0 = 1,$$
 denoted $\GQ^{q,p}_s$, is defined as follows. Let $\xx =\sbr*{x_1, x_2, \hdots, x_n}\in \R^{nd}$. Let $y_i \eqdef |x_i|/\|\xx\|_{q,p} \in \R^d$ for all $i\in [1,...,n]$.
Then
\begin{equation}
\label{eq:glob_quant}
  \GQ^{q,p}_s(\xx) \eqdef \norm{\xx}_{q,p}  \frac{1}{n}\sum_{i=1}^n \signum(x_i) \circ \xi_i(y_i) \; , \
\end{equation}
where $\xi_i(y_i)$ is an independent element-wise random rounding operator such that
\begin{equation}
\label{eq:rounding}
\rbr*{\xi_i(y_i)}_j \eqdef 
\begin{cases} 
     l_{u^j_i} & \mbox{with probability } \frac{\rbr*{y_i}_j-l_{u^j_i+1}}{l_{u^j_i}-l_{u^j_i+1}} \\ 
     l_{u^j_i+1} & \mbox{otherwise} \end{cases},
\end{equation}
for $j \in [d]$,
where $u^j_i \in \{0,1,2, \hdots, s\}$ and $  l_{u^j_i+1} \leq \rbr*{y_i}_j \leq l_{u^j_i}$. 
\label{def:gqsgd}
\end{definition}

\subsection{Linear and Exponential Dithering}
\label{sec:exp_design}
We consider two approaches to partitioning the quantization levels $l_{i}$ in Global-QSGD:
\begin{itemize}
    \item \textit{Linear Dithering} (\( \GL^{q,p}_s \)): The straightforward way is to divide the interval $[0, 1]$ into equal partitions as \( l_{i} = \nicefrac{s-i}{s} \). However, as gradients converge to zero during training, this method becomes less accurate since most values fall into the same levels.
    \item \textit{Exponential Dithering} (\( \GE^{q,p}_s \)): To provide higher precision for near-zero values, we propose a nonuniform interval partitioning approach. Inspired by \citet{horvath2019natural}, we divide intervals exponentially as \( l_{s} = 0 \) and \( l_{i} = \nicefrac{1}{2^{s-i}} \). We use base \( 2 \) for natural compatibility with IEEE standard floating point representation.
\end{itemize}

Supporting 255 levels, we adopt 8-bit as the default quantization bit-width, since most hardware accelerators natively support this numerical representation.
For bit-widths without direct hardware support, a manual implementation of bit-wise operations is necessary.
To demonstrate that Global-QSGD is theoretically compatible with any bit precision, we also implement the 4-bit version on the NVIDIA Ampere architecture; results are presented in \S~\ref{sec:evaluation}.

The implementation of \(\GL^{q,p}_s\) is simple and inherently compatible with Allreduce, since mapping to uniform intervals is a homomorphic operation, allowing efficient aggregation by summing the quantized integers values directly.
However, for \(\GE^{q,p}_s\), summation via Allreduce becomes more complex, as the quantized values follow an exponential form \( 2^k \). To ensure compatibility with Allreduce, we introduce stochastic unbiased exponential rounding, denoted as \( \NC \), following the notation in \citet{horvath2019natural}. This rounding scheme introduces a minor variance increase per step, specifically a factor of \( \nicefrac{9}{8} \). Over multiple aggregation steps, this accounts for $\rbr*{\nicefrac{9}{8}}^{\text{\# aggregation steps}}$, i.e., $\rbr*{\nicefrac{9}{8}}^{\log(n)} \leq n^{0.17}$ for Tree-Allreduce, resulting in a small increase in variance—for example, only 1.6× for \( n=16 \) and 3.25× for \( n=1024 \).

Finally, we discuss a strong advantage of \( \GE^{q,p}_s \)'s scaling properties with the number of nodes $n$, due to better integer overflow resistance.
Let us look at the following example, where integer values are represented with $A$ bits ($A \in \N$). In the case of linear dithering, the maximum aggregated value is $ns$. Besides, one bit is needed to hold the sign. Therefore, we require $1 + \log(s + 1) + \log(n) \leq A$, which cannot be satisfied for any $s$, for example in the case with $n=16$ and $A=4$. On the other hand, the maximum aggregated value with exponential dithering is $n2^s$; but since we only communicate the exponent, we obtain an improved scaling, as the maximum integer communicated is $s + \log(n)$. Therefore, we only require $1 + \log(s + 1 + \log(n)) \leq A$, which is satisfied for $s \leq 3$.  This is because exponential dithering scales as $\log(\log(n))$ with $n$ instead of just $\log(n)$.

\begin{algorithm}[tb]
    \begin{algorithmic}[1]
    \REQUIRE $(\sign_1, e_1), (\sign_2, e_2), s$
    \STATE $k = -\floor{\log\rbr*{(2^{-m} + \sbr*{p - 2^{-m}}_+)}}$
    \STATE $\eonenotzero = \mathbf{1}(e_1 > 0)$ \COMMENT{whether $e_1$ is nonzero}
    \STATE $\etwonotzero = \mathbf{1}(e_2 > 0)$ \COMMENT{whether $e_2$ is nonzero}
    \STATE $\sign_{12}=\sign_1 \cdot \sign_2 \cdot \eonenotzero \cdot \etwonotzero$
    \STATE $\diff = \abs{e_1 - e_2} - (1 - \sign_{12}) // 2$
    \STATE $\smaller = (\mathbf{1}(e_1 \leq e_2) + !\etwonotzero) \cdot \eonenotzero$
    \STATE $\nonz = 1 - \mathbf{1}\rbr*{e_1=e_2 \land \sign_{12} = -1}$
    \STATE $\sign_{\text{result}} = \sign_1 \cdot \smaller + \sign_2 \cdot (1 - \smaller)$
    \STATE $e_{\text{result}} = (e_1 \cdot \smaller + e_2 \cdot (1 - \smaller)) \cdot \nonz$ \\
    \hspace{1.5em} $- \sign_{12} \mathbf{1}(k>\diff) \cdot \nonz$
    \RETURN $\sign_{\text{result}},  e_{\text{result}}$
    \end{algorithmic}
    \caption{Reduce Function for $\GE^{q,p}_s$}
    \label{alg:reduce_exp}
\end{algorithm}

\subsection{Reduce Function for Exponential Dithering}
\label{sec:reduce}
To efficiently implement \( \GE^{q,p}_s \), we introduce the \textit{exponential reduce} function (Algorithm~\ref{alg:reduce_exp}), which aggregates values using integer-based arithmetic.
This GPU-friendly integer-based and branch-free algorithm ensures efficient parallelization with CUDA.

Since our reduction function acts element-wise, we only consider here a one-dimensional case. 
Before quantization and aggregation, we divide all values by $2n$, where $n$ is the number of nodes, to ensure that the maximum power of two we encounter during the aggregation is $-1$ that corresponds to $\nicefrac{1}{2} = 2^{-1}$. Therefore, all the encountered exponents during the aggregation are guaranteed to be negative. In this way, we can represent zero as 0 instead of $2^0$.

Next, we define the representation used during the aggregation by our exponential reduce function, which assumes exponential dithering is locally executed prior to its invocation.
Let $\sign \in \cbr{-1,+1}$\footnote{Zeros have any sign.} be the sign and $e \in \cbr*{\N \cup \cbr{0}}$ be the communicated nonnegative integer-valued exponents. Then, then real number $x$ that corresponds to the pair $(\sign, e)$ is defined as 
\begin{align}
    \label{eq:dithering_representation}
     x = 
\begin{cases}
      0, & \text{if } e=0, \\
      \sign 2^{-e}, & \text{otherwise}.
\end{cases}
\end{align}

We proceed with the derivation for the reduce function. Let $x_1, x_2 \in \R$ represented by $(\sign_1, e_1), (\sign_2, e_2)$, respectively, be the values to be summed using the reduce function.
To facilitate efficient rounding, define the maximum difference $\abs{e_1 - e_2}$ that can appear during aggregation as
$$
k = -\floor{\log\rbr*{2^{-m} + \sbr*{p - 2^{-m}}_+}},
$$
where $p \sim \text{Unif}\sbr{0, 1}$ is a sample of the uniform distribution in the interval $[0, 1]$,\footnote{Note this can be precomputed.} $m \eqdef s + 1$, and $\sbr*{x}_+ \eqdef \max\cbr{0, x}$. It holds
$$
\Prob\rbr*{k > b} = \Prob\rbr*{-k < -b} =\Prob\rbr*{p < 2^{-b}} = 2^{-b}.
$$
Note that the support set for $k$ is $\cbr*{0, 1, \hdots, m}$, and for $b \in \cbr*{0, 1, \hdots, m-1}$.
Without loss of generality, we assume that $\sign_1 = 1$ and $0 < e_1 \leq e_2$. We later discuss the case with $e_1=0$ or $e_2=0$.
If $\sign_2 = 1$, then
\begin{align*}
\NC\rbr*{2^{-e_1} + 2^{-e_2}} = 
\begin{cases}
    2^{-e_1 + 1}, & \text{w.p. } 2^{e_1 - e_2}, \\
    2^{-e_1}, & \text{w.p. } 1 - 2^{e_1 - e_2},
\end{cases}
\end{align*}
where w.p. stands for ``with probability.'' We note that 
$$
p < 2^{e_1 - e_2} \iff k > e_2 - e_1.
$$
Therefore, 
\begin{align*}
\NC\rbr*{2^{-e_1} + 2^{-e_2}} = 
\begin{cases}
      2^{-e_1 + 1}, & \text{if } k > e_2 - e_1, \\
      2^{-e_1}, & \text{otherwise}.
\end{cases}
\end{align*}
Analogously, when $\sign_2 = -1$, we write $\NC\rbr*{2^{-e_1} - 2^{-e_2}}$ as:
\begin{align*}
\begin{cases}
      0, & \text{if } e_1 = e_2 \\
      \begin{cases}
      2^{-e_1 - 1}, & \text{w.p. } 2^{e_2 - e_1 - 1}, \\
      2^{-e_1}, & \text{w.p. } 1 - 2^{e_2 - e_1 - 1}, 
\end{cases} & \text{otherwise}.  
\end{cases}
\end{align*}
Equivalently, we can write $\NC\rbr*{2^{-e_1} - 2^{-e_2}}$ as:
\begin{align*}
\begin{cases}
      0, & \text{if } e_1 = e_2 \\
      \begin{cases}
      2^{-e_1 - 1}, & \text{if } k > e_2 - e_1 - 1,  \\
      2^{-e_1}, & \text{if } k \leq e_2 - e_1 - 1,
\end{cases} & \text{otherwise}.  
\end{cases}
\end{align*}
In general, we compare $k$ to the following quantity \begin{align*}
\diff = \abs{e_1 - e_2} - (1 - \sign_{12}) // 2,
\end{align*}
where $\sign_{12} \eqdef \sign_1  \sign_2  \in \cbr{-1, 1}$ and $//$ corresponds to the integer division. It is easy to check that the above quantity recovers all the above-mentioned cases. To determine which exponent is smaller, we use $\smaller = \mathbf{1}(e_1 \leq e_2) \in \cbr{0, 1}$, where $\mathbf{1}$ is the indicator function.  Finally, to filter out the case of different signs and the same exponent, we use $\nonz \eqdef 1 - \mathbf{1}\rbr*{e_1=e_2 \mbox{ and } \sign_{12} = -1}  \in \cbr{0, 1}$. Then, we obtain the resulting sign and exponent as 
\begin{align*}
    \sign_{\text{result}} &= \sign_1 \cdot \smaller + \sign_2 \cdot (1 - \smaller) \\
    e_{\text{result}} &= \rbr*{e_1 \cdot \smaller + e_2 \cdot (1 - \smaller)} \cdot \nonz \\
    &\quad - \sign_{12} \cdot \mathbf{1}(k>\diff) \cdot \nonz.
\end{align*}
To incorporate zeros, we first define the following variables to identify is an exponent is greater than zero
\begin{align*}
    \eonenotzero &= \mathbf{1}(e_1 > 0) \\
    \etwonotzero &= \mathbf{1}(e_2 > 0).
\end{align*}
In the case where at least one of the exponents is zero, we do not change the exponents by adding or subtracting zero. This can be achieved by redefining $\sign_{12} \in \cbr{-1, 0, 1}$ to 
\begin{align*}
    \sign_{12} \eqdef \sign_1 \cdot \sign_2 \cdot \eonenotzero  \cdot \etwonotzero .
\end{align*} 
Furthermore, we need to redefine $\smaller$ to account for zeros. This can be achieved by
\begin{align*}
    \smaller \eqdef (\mathbf{1}(e_1 \leq e_2) + !\etwonotzero)  \cdot \eonenotzero,
\end{align*}
where ! denotes logical negation.
This concludes our construction of the reduce function for the exponential dithering.

\section{Variance Analysis}
\label{sec:variance}
We provide the variance analysis of Global-QSGD, which is the key ingredient to ensure convergence in \S~\ref{sec:convergence}.
The theoretical framework extends from~\citep{cordonnier2018convex, Mishchenko27092024, koloskova2019decentralized, stich2018sparsified}.
In \S~\ref{sec:unbias}, we extend the concept of \textit{Unbiased Compressors} ($\U^{d}(\omega)$) to a broader class, \textit{Unbiased Distributed Mean Compressors} ($\U^{n, d}(\theta)$). We then prove that $\GQ^{q,p}_s \in \U^{n, d}(\theta)$, establishing its unbiasedness.  
Building upon this property, in \S~\ref{sec:var-bound} we demonstrate that $\GQ^{q,p}_s$ has bounded variance, a crucial condition to ensure convergence.
\ifarxiv
The proofs of all the theorems and lemmas are in Appendix~\ref{app:proofs}.
\else
The proofs of all the theorems and lemmas are in~\citep{xin2025globalqsgd}[Appendix~\ref{app:proofs}].
\fi
\subsection{Unbiasedness}
\label{sec:unbias}
We start by defining the Unbiased Compressor as $\U^{d}(\omega)$:
\begin{definition}[Unbiased Compressor]
\label{def:omegaquant} A randomized mapping $\cC\colon \R^d \to \R^d$  is an {\em unbiased compressor}  if there exists $\omega \geq 0$ such that $\forall x \in \R^d:$
\begin{equation}
\label{eq:omega_quant}
 \E{\cC(x)}=x, \qquad \E{\norm{\cC(x) - x}_2^2} \leq \omega \norm{x}_2^2.
 \end{equation}
If this holds, for simplicity we will write $\cC\in \U^d(\omega)$.
\end{definition}
\noindent We now generalize this notion to distributed settings as follows:
\begin{definition}[Unbiased Distributed Mean Compressor]
\label{def:thetaquant}
For any $x_1, x_2, \hdots, x_n \in \R^d$, let 
\begin{equation}
\label{eq:x_global}
\xx \eqdef \sbr*{x_1, x_2, \hdots, x_n}  \in \R^{nd}, \qquad \bx \eqdef \frac{1}{n}\sum_{i=1}^n x_i \;.
\end{equation}
A randomized mapping $\cG\colon \R^{nd} \to \R^d$  is an {\em unbiased distributed mean compressor} if there exists $\theta \geq 0$ such that $\forall \xx \in \R^{nd}:$
\begin{equation}
    \label{eq:theta_quant}
     \E{\cG(\xx)}=\bx, \qquad \E{\norm{\cG(\xx) - \bx}_2^2} \leq  \frac{\theta}{n} \norm{\xx}_{2,2}^2. 
 \end{equation}
If this holds, for simplicity we will write $\cG\in \U^{n, d}(\theta)$.
\end{definition}
To show that Definition~\ref{def:thetaquant} is more general than Definition~\ref{def:omegaquant}, we formalize the following lemma:
\begin{lemma}[$\C \subset \U^{n, d}$]
\label{lem:c_subset_d}
If $\cC_1, \cC_2, \hdots, \cC_n  \in \U^n(\omega)$ and they are independent, then $\cG\colon \R^{nd} \to \R^d$ defined as Equation \ref{eq:g(x)} and belongs to $\U^{n, d}\rbr*{\nicefrac{\omega}{n}}$.
\begin{equation}
    \label{eq:g(x)}
    \cG(\xx) \eqdef \frac{1}{n}\sum_{i=1}^n\cC_i(x_i).
\end{equation}
\end{lemma}

In the next lemma, we show that $\GQ^{q,p}_s \in \U^{n, d}(\theta)$ has an interesting reduction property that helps us to analyze its theoretical properties using known results for the case $n=1$~\citep{qsgd2017neurips, horvath2019natural}.
\begin{lemma}
\label{lem:global_quant_reduction}
Let $\cQ^{q,p}_s(\xx)\eqdef \norm{\xx}_{q,p} \signum(\xx) \circ \xi(\yy)$, where 
$\xi(\yy) = \sbr*{\xi_1(y_1), \xi_2(y_2), \hdots, \xi_n(y_n)}$ and $\xi_i(y_i)$ is defined in \eqref{eq:rounding}. Then, 
\begin{align}
    \label{eq:reduction}
    \GQ^{q,p}_s(\xx)= \frac{1}{n} \sum_{i=1}^n \rbr*{\cQ^{q,p}_s(\xx)}_i,
\end{align}
where $\rbr*{\cQ^{q,p}_s(\xx)}_i$ refers to coordinates $[(i-1)d + 1, \hdots, id]$.
Moreover, if $\cQ^{q,p}_s \in \cC(\omega)$ then $\GQ^{q,p}_s \in \U^{n, d}(\theta)$ with $\theta = \nicefrac{\omega}{n}$.
\end{lemma}
Note that there is a difference in the dependence on the dimension, i.e., $d \to nd$, since we work with the concatenated vector $\xx$.
\subsection{Variance Bound}
\label{sec:var-bound}
Next, we derive the exact bound on the variance of  $\GL^{q,p}_s$ and  $\GE^{q,p}_s$. In addition, for the special case of $p=q=2$, we establish an upper bound on sparsity, i.e., the sum of zero norms of the communicated vectors ($\norm{y}_0$ denotes the number of non-zero elements of $y$).
\begin{theorem}
\label{thm:sparsity}
If $p, q \geq 2$ then $\GL^{q,p}_s \in \U^{n, d}\rbr*{\frac{\sqrt{d}}{\sqrt{n}s}}$ for $s \leq \sqrt{nd}$, and $\GE^{q,p}_s \in \U^{n, d}\rbr*{\frac{1}{8n} + \frac{\sqrt{d}}{\sqrt{n}2^{s-1}}}$ for $s \leq 1 + \log\rbr*{\sqrt{nd}}$. Moreover, if $p=q=2$ then for any $\xx \in \R^{nd}$
\begin{align*}
    \sum_{i=1}^n \norm*{\rbr*{\cL^{2,2}_s(\xx)}_i}_0 \leq s^2 + \sqrt{nd}, \\
    \sum_{i=1}^n \norm*{\rbr*{\cE^{2,2}_s(\xx)}_i}_0 \leq 2^{2s - 2} + \sqrt{nd}.
\end{align*}
\end{theorem}
%
%
The above theorem guarantees that we can achieve $\cO(\sqrt{nd})$ compression ratio for $s=\cO(1)$. 
Furthermore, we note that the variance bound scales better for exponential dithering with the number of levels $s$, which means that exponential dithering exhibits a smaller relative compression error for larger $s$.
Using these definitions, standard convergence analysis (e.g., \citep{gorbunov2020unified}) extends naturally to our distributed setting. In particular, employing an unbiased distributed mean compressor $\cQ\in\U^{n,d}(\theta)$ increases the complexity of the iteration by a factor of $1+\theta n$ (recovering $1+\omega$ for standard compressors). 

\section{Convergence Analysis}
\label{sec:convergence}
We show how the standard analysis for unbiased compressors can be generalized to the proposed unbiased distributed mean compressor. Since we construct compressors to be unbiased, the only extra challenge of the analysis is to bound the variance of the gradient estimator.\footnote{The same applies to the analysis of non-smooth functions, where gradients are replaced with subgradients.}
\begin{lemma}[Variance bound]
    \label{lem:varianceBound}
    Let $x \in \R^d$ be deterministic and $\EE{\xi_i \sim \cD_i}{\nabla f_i(x, \xi_i)} = \nabla f_i(x)$ for all $i \in [n]$. Then
    \begin{align}
    \label{eq:variance_bound}
        &\E{\norm*{\cQ\rbr*{\sbr{\nabla f_1(x, \xi_1), \ldots, \nabla f_n(x, \xi_n)}} - \nabla f(x)}_2^2} \leq \notag \\
        &\qquad \frac{\rho}{n}\sum_{i=1} \norm*{\nabla f_i(x))}_2^2 \qquad + \\ \notag
        &\rbr*{\rho + \frac{1}{n}} \frac{1}{n}\sum_{i=1}\EE{\xi_i \sim \cD_i}{\norm*{\nabla f_i(x, \xi_i) - \nabla f_i(x)}_2^2} 
    \end{align}
    holds with $\rho = \frac{\omega}{n}$ for $\cQ\rbr*{\sbr{\nabla f_1(x, \xi_1), \ldots, \nabla f_n(x, \xi_n)}} \eqdef \frac{1}{n}\sum_{i=1} \cC_i(\nabla f_i(x, \xi_i))$, where $\cC_1, \cC_2, \hdots, \cC_n  \in \U^n(\omega)$ are independent. Furthermore, if $\cQ \in \U^{n, d}(\theta)$ then \eqref{eq:variance_bound} holds with $\rho = \theta.$
\end{lemma}
 In Lemma \ref{lem:varianceBound}, we show that applying the unbiased distributed mean compressor $\cQ \in \U^{n, d}(\theta)$ yields the same bound on variance\footnote{We assume that the noise due to sampling and compression are independent.} as the difference that $\nicefrac{\omega}{n}$ is replaced with $\theta$.
The rest of the analysis is the same as the general analysis of \citet{gorbunov2020unified}, which shows that the worst-case increase in the number of iterations to achieve the same precision as the algorithm without compression is by the factor $1 + \omega$ that translates to $1 + \theta n$ for the unbiased distributed mean compressors $\cQ \in \U^{n, d}(\theta)$ \citep[Theorem 4.1 for Alg. 1: SGD and Alg. 13: Quantized-SGD]{gorbunov2020unified}. For exponential dithering, this would be a factor of $1 + n^{1.17}\rbr*{\nicefrac{1}{8n} + \nicefrac{\sqrt{d}}{\sqrt{n}2^{s-1}}}$. In this estimate, we consider all the sources of variance for Allreduce, i.e., variance due to initial quantization and extra variance due to stochastic rounding during Allreduce.

For example, let us consider the setup where we use exponential dithering to decrease communication precision from $32$ to $8$ bits, the dimensionality of the model is $d=10^6$, and the number of machines is $n=16$. We have to reserve $1$ bit for the sign and the other $7$ bits for levels. Plugging these numbers into our formula yields the worst-case increment in the expected number of iterations, which is $20\%$, while we save $75\%$ of communication. Therefore, if communication is a significant bottleneck, our quantization leads to a guaranteed speed-up. Note that this is only the worst case, and in real-world scenarios, this can be much less, e.g., the empirical compression error in our experiments is below $0.5\%$ for exponential dithering.

Furthermore, while our analysis primarily focuses on SGD, the concept of global quantization is intentionally designed to be broadly compatible with various optimizers, including popular ones such as Nesterov momentum and Adam. This compatibility stems from its core property of being unbiased with bounded variance, aligning it with the theoretical foundations underlying these optimizers. Typically, optimizers such as SGD are predicated on the use of stochastic gradients that are unbiased and maintain variance within specific limits—conditions supported by our method. For example, integration with adaptive methods can be directly obtained by generalizing the results of \citet{defossez2022a}.
Our experiments with Transformer-XL using Adam confirm this compatibility.

In summary, global quantization is compatible with standard optimization methods designed for efficient distributed learning, such as distributed variants of Adam and SGD. By maintaining unbiased gradient estimates with bounded variance, our quantization method preserves convergence guarantees comparable to those of uncompressed methods while significantly reducing communication costs.

\section{Performance Model}\label{sec:Communication Modeling}
Gradient compression can only be beneficial if the introduced computation overhead can be compensated by communication gains. Thus, we propose a performance model to analyze how Global-QSGD can speed up training.

In practice, there are two popular Allreduce implementations: Ring-based and Tree-based. We adopt Tree-based Allreduce for our evaluation, as it has fewer reduction steps in our 4-GPU setup, where each quantized reduction introduces computation overhead and accuracy loss due to random rounding.
The tree Allreduce algorithm first recursively aggregates the gradients, then does a recursive-doubling Allgather. The depth of the tree is $2\log(n)$ where $n$ is the number of workers.
The performance of Allreduce is well studied \citep{Allreduce} and it is commonly modeled as $2\log(n)\alpha + 2\frac{\log(n)S}{\beta} + \frac{\log(n)S}{\gamma}$,  where $\alpha$ is the propagation delay in seconds, $S$ is the size of gradients in bytes, $\beta$ is the bandwidth (byte/s), and $\gamma$ is the computation speed (byte/s).
The first term represents the propagation delay, the second term is the bidirectional transmission delay, and the last term is the computation cost.

We denote the quantized gradient size as $\hat{S}$, and the computation cost with custom reduction as $\hat{\gamma}$. The bandwidth and propagation delay remain the same. Denote the quantization and dequantization time as $\delta$ factor of the gradient size. Then the Allreduce performance for Global-QSGD is $2\log(n)\alpha + 2\frac{\log(n)\hat{S}}{\beta} + \frac{\log(n)\hat{S}}{\hat{\gamma}} + {\delta}S$.

We denote the quantization ratio $\rho = \frac{S}{\hat{S}}$ ( by default $\rho=4$).
We consider the computation cost negligible ($\delta=0$), as the quantization and dequantization operations are executed only once per Allreduce invocation, which ideally can be overlapped with the communication if the model is communication-bottlenecked. Therefore, the cost of the reduction operation will dominate the performance as the number of workers increases.
We denote the computation overhead as $\omega =  \frac{\gamma}{\hat{\gamma}}$.
We empirically measure\footnote{Measured on a A100 with 25 MB data (default bucket size in PyTorch).} this overhead to be $\omega_{\cL} = 1$ and $\omega_{\cE} = 79$, for linear and exponential dithering, respectively. The reduction operation of linear dithering is the native arithmetic summation, which has a similar time for uint8 and float32 data types in our experiments. Instead, the custom reduction used for exponential dithering involves more arithmetic operations, which yield a higher computation overhead.

The condition of performance gain is that per-batch training time after applying Global-QSGD is less than the per-batch training time without quantization, which is:
\begin{align*}
            2\log(N)\alpha + 2\frac{\log(N)S}{\beta} &+ \frac{\log(N)S}{\gamma} > \displaybreak[1]\\
            2\log(N)\alpha + 2\frac{\log(N)\hat{S}}{\beta} &+ \frac{\log(N)\hat{S}}{\hat{\gamma}} + {\delta}S \displaybreak[1]
\end{align*}

\ifarxiv
Thus, the condition for Global-QSGD to speed up training is as follows (details of the derivation are in Appendix~\ref{app:modeling}):
\else
Thus, the condition for Global-QSGD to speed up training is as follows (details of the derivation are in~\citep{xin2025globalqsgd}[Appendix~\ref{app:modeling}]):
\fi

\begin{equation}
\label{eq:condition}
\left\{ 
\begin{array}{ll}
\beta > \frac{6\gamma}{(\omega - 4)}, & \text{if } (\omega < 4),\\
\beta < \frac{6\gamma}{(\omega - 4)}, & \text{if } (\omega > 4).
\end{array} 
\right.
\end{equation}
With $\omega_{\cL} = 1$, theoretically linear dithering is guaranteed to speed up training (since $\beta > 0$). In the case of exponential dithering ($\omega_{\cE} = 79$), there is a training speed up if the relation $\beta < 0.08\gamma$ holds. Our evaluation is conducted using A100 GPU ($\gamma=2$ TB/s)  with P2P ($\beta=53.9$ GB/s) or SHM ($\beta=5.4$ GB/s). Therefore, both the P2P- and SHM-based communication satisfy the speedup condition.
\begin{figure*}[t!]
    \centering
    \begin{subfigure}{.31\textwidth}
        \centering
        \includegraphics[width=1\textwidth]{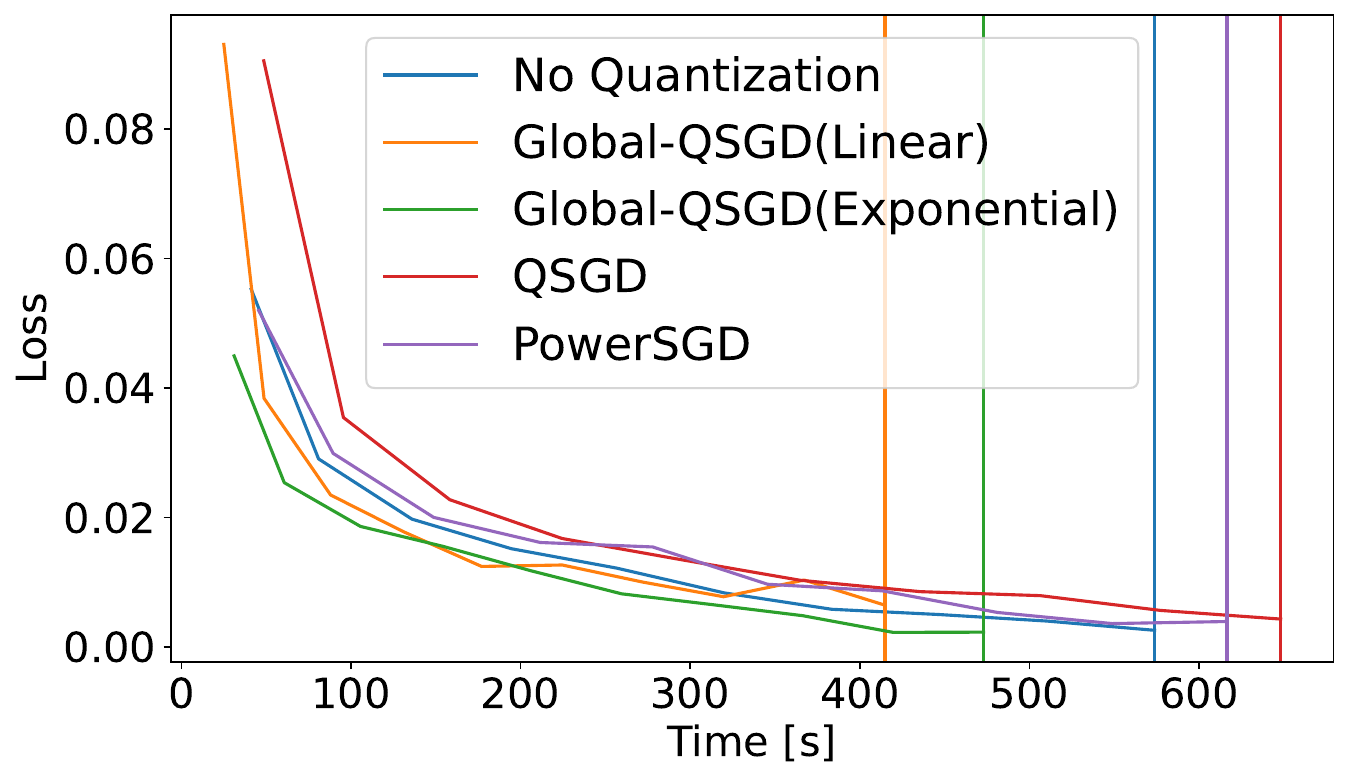}
        \caption{DeepLight}
        \label{fig:DeepLight}
    \end{subfigure}%
    \hfill  
    \begin{subfigure}{.31\textwidth}
        \centering
        \includegraphics[width=1\textwidth]{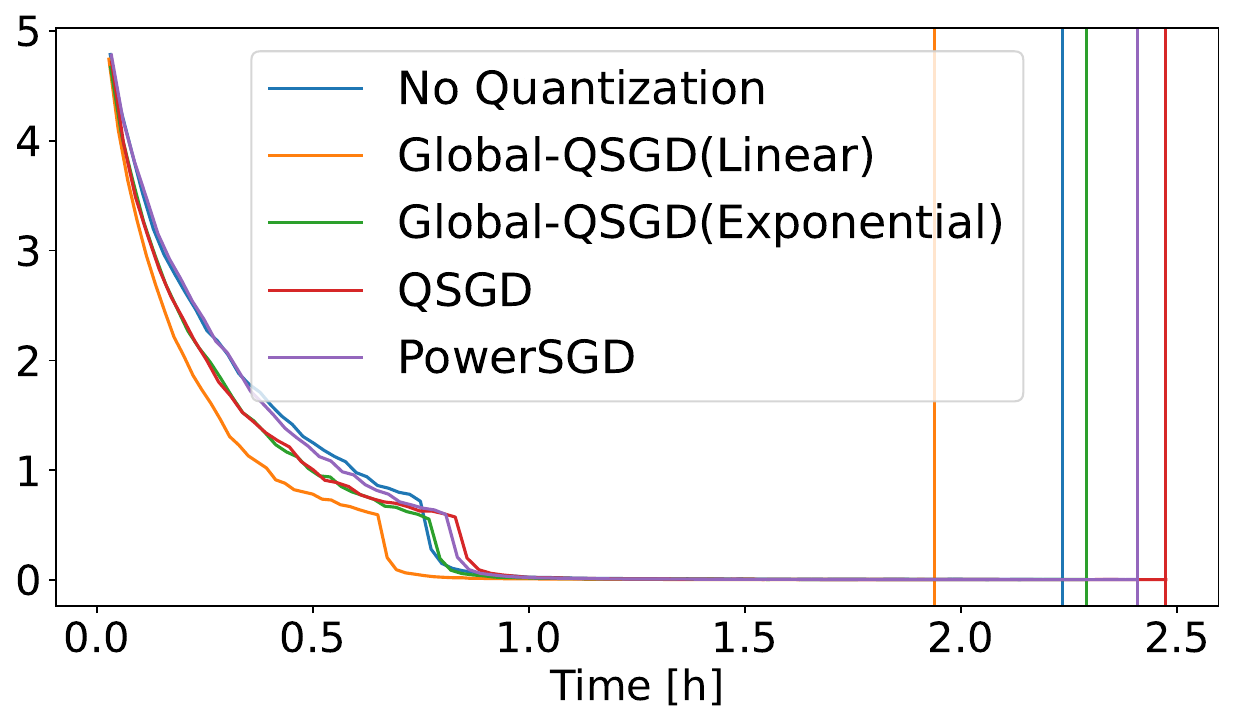}
        \caption{Wide ResNet-101-2}
        \label{fig:ResNet101}
    \end{subfigure}%
    \hfill  
    \begin{subfigure}{.31\textwidth}
        \centering
        \includegraphics[width=1\textwidth]{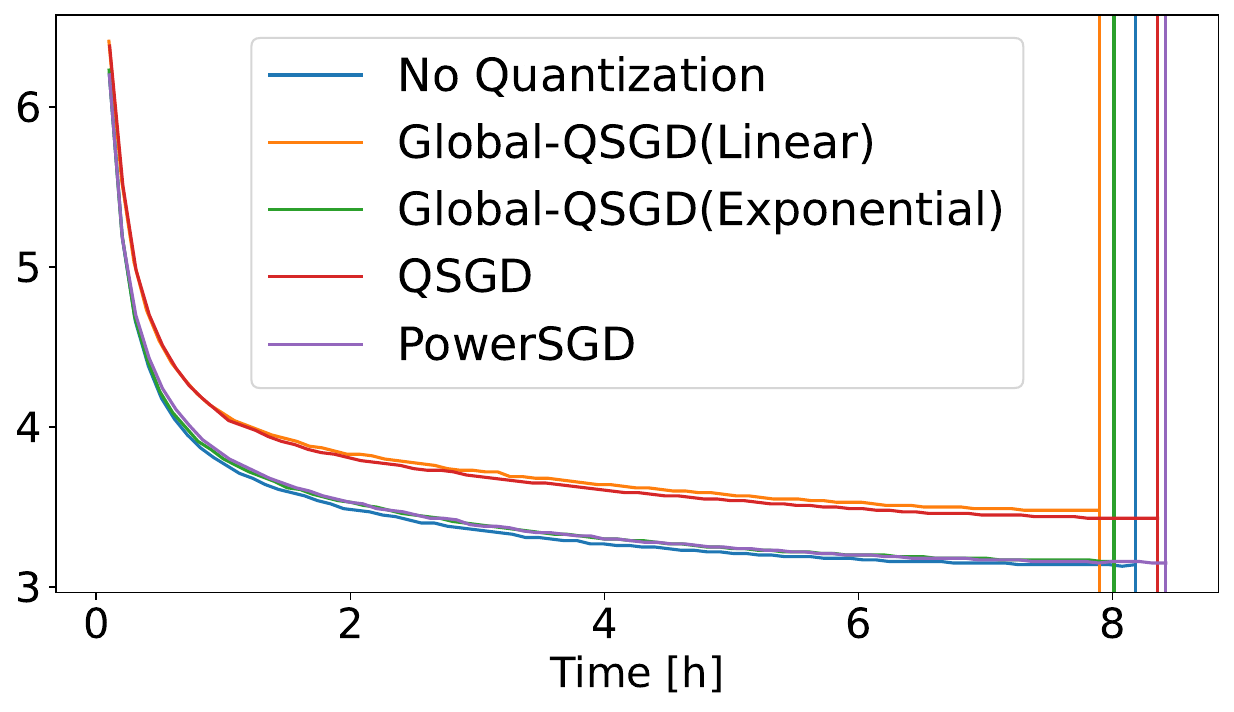}
        \caption{TransformerXL}
        \label{fig:TransformerXL}
    \end{subfigure}
    \vspace{12pt}
    \caption{Training loss in P2P. Vertical lines represent the completion times after a fixed iteration count.}
    \label{fig:Loss}
    \vspace{1em}
\end{figure*}

\section{Implementation}
\begin{figure}[t!]
    \centering
    \includegraphics[width=0.83
    \linewidth]{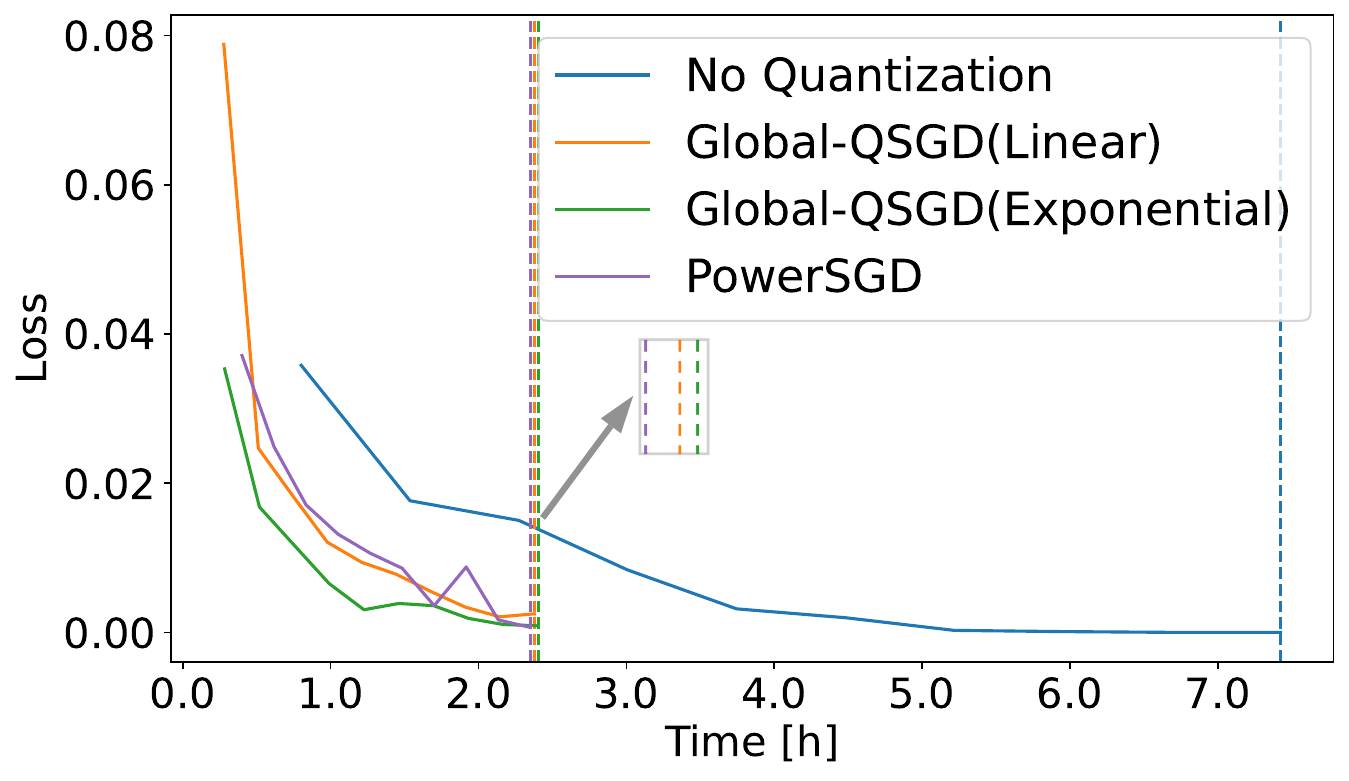}
    \caption{DeepLight training loss in GCP with 64 nodes.}
    \label{fig:DeepLightCloud}
    \vspace{1.5em}
\end{figure}
        
        

\begin{figure}[t!]
    \centering
    \includegraphics[width=0.4\textwidth]{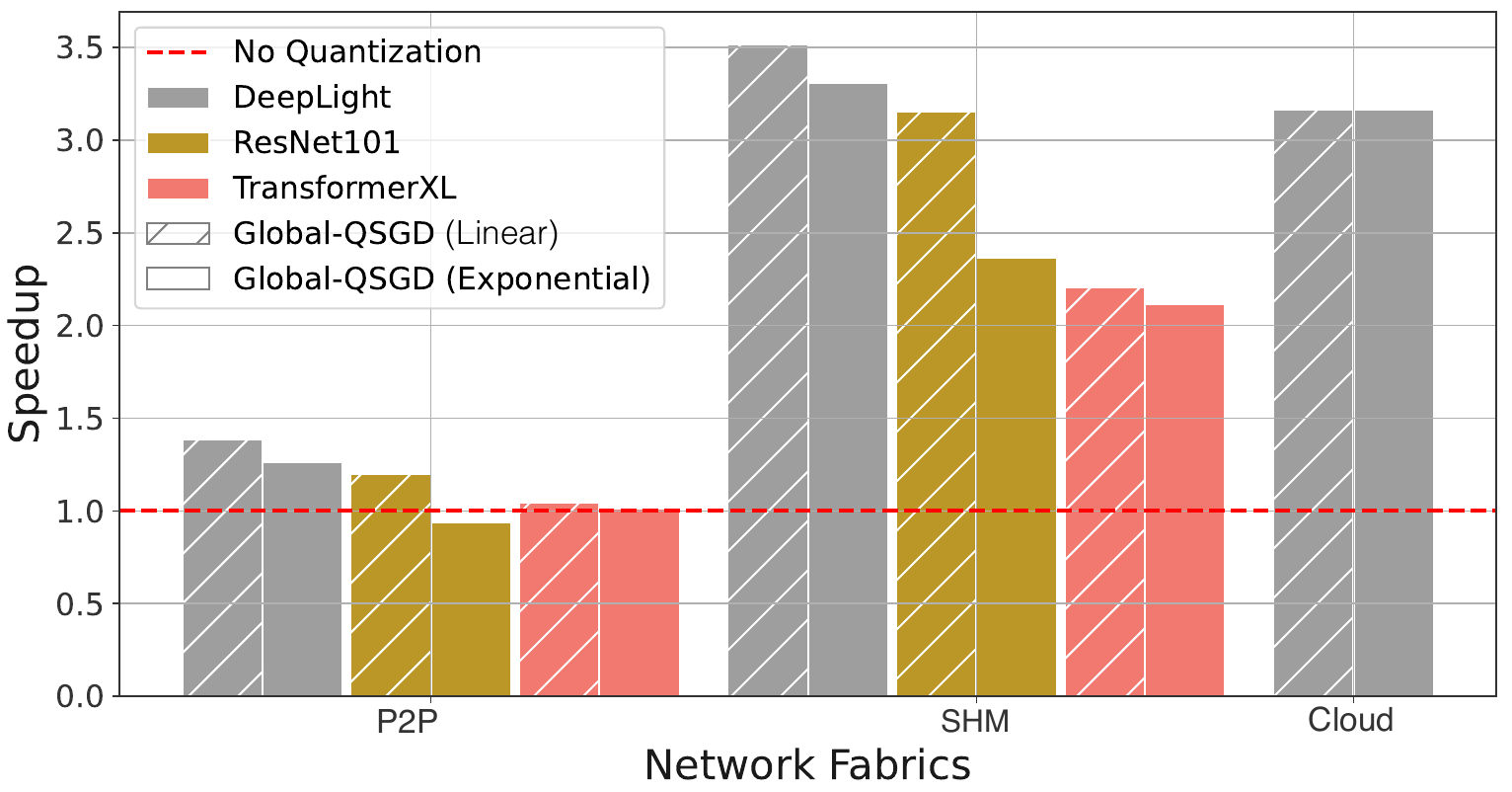}
    \caption{Training throughput speedup normalized to no quantization.}
    \label{fig:speedup}
    \vspace{2em}
\end{figure}
Our implementation of Global-QSGD supports both linear ($\GL^{q,p}_s$) and exponential ($\GE^{q,p}_s$) dithering methods, with default configuration using the $\ell_{inf}$-norm ($p=q=\infty$) and 8-bit precision ($s = 255$).
For ultra-low precision scenarios (4-bit or lower), our implementation can be seamlessly extended without modification when hardware support exists (such as on Hopper GPUs with native 4-bit operations).

In the absence of native hardware support, a custom bit-wise operation is required.
We generally encourage users to use the bit-width supported by the hardware, i.e., 8-bit on Ampere and 4-bit on Hopper.
We also implemented a 4-bit version $\GL^{q,p}_s$ on Ampere GPUs,\footnote{This implementation idea can be naturally extended to $\GE^{q,p}_s$.} where we manually pack 4-bit values into 8-bit.
Note that this implementation is used to demonstrate Global-QSGD's effectiveness in convergence on low-bit. As this is not hardware- nor CUDA-optimized, its performance is not representative.

As discussed in \S~\ref{sec:exp_design}, linear dithering ($\GL^{q,p}_s$) encounters significant limitations in low-bit large-scale scenarios.
For example, with 4-bit quantization, after allocating 1 bit for the sign and 2 bits to handle potential overflow across 4 workers, only a single bit remains for actual gradient values. This severely restricts precision and cannot be scaled further.
To address this limitation, we put the averaging approach inside each Allreduce step rather than accumulating sums until the final step.
While this introduces some precision loss due to rounding operations at each averaging step, our experiments demonstrate that the method still maintains fair convergence properties.

The algorithm is wrapped in a custom Allreduce module integrated into PyTorch as a DDP hook.
Our hook procedure is invoked by the granularity of a gradient bucket, which by default has a size of at least 25 MB.
The procedure consists of three steps: Quantization, Allreduce, and Dequantization.
In the case of exponential dithering, the Allreduce step uses a custom reduction function.
To achieve this, we implement a customized Allreduce algorithm using NCCL's point-to-point asynchronous communication API. We support both ring and tree Allreduce.
We develop the quantization, de-quantization, and the custom reduction for exponential reduction in CUDA to optimize GPU performance.

To use Global-QSGD, the user simply has to register our DDP hook  with one line of code like 
\textsf{model.register\_comm\_hook()}.

\section{Evaluation}\label{sec:evaluation}
The aim of our evaluation is to illustrate that our proposal is practical and beneficial in a range of scenarios, including with different bandwidths among GPUs. We focus on measuring the speedup of training throughput in three distinct domains of DL applications. Through task-specific metrics on the test set, we show that Global-QSGD does not impair the model's generalization ability.
We further illustrate that, compared to linear dithering, exponential dithering achieves better overall performance due to its smaller error for small values, despite its additional overhead from the exponential reduction function.

\smartparagraph{Setup.}
Our experiments are primarily run on one ASUS ESC N4A-E11 server that runs Ubuntu 22.04 with CUDA 11.6, and we use PyTorch 1.13.0.
The server is equipped with 4 NVIDIA A100 GPUs, each with 40 GB of RAM. GPUs are peer-to-peer connected by 4 NVLink channels (4th generation).
To tease out the effects of bandwidth on training speed, we run experiments with two interconnects: P2P, which employs NVIDIA GPU Direct allowing data to transmit via NVLink directly without the interference of CPU and host memory; SHM, which uses the host memory as a middle buffer; thus, the data will transmit through PCIe.
The empirically measured uni-directional bandwidth for 25 MB data is 53.9 GB/s and 5.4 GB/s for P2P and SHM, respectively.
\ifarxiv
Details for bandwidth measurements are in Appendix~\ref{app:bandwidth}.
\else
Details for bandwidth measurements are in~\citep{xin2025globalqsgd}[Appendix~\ref{app:bandwidth}].
\fi

\begin{table}[tb]
    \centering
    \caption{Summary of benchmarks used in this work.}
    \label{tab:benchmark}
    \vspace{6pt}
    \resizebox{0.45\textwidth}{!}{
    \begin{tabular}{|l|l|c|c|}
    \hline
        Model & Dataset & Parameter Size & Training Epochs \\ \hline
        DeepLight & Tiny Criteo & 607,959,381 & 10 \\ \hline
        Wide ResNet-101-2 & MiniImageNet & 126,886,696 & 90 \\ \hline
        TransformerXL & WikiText-103 & 191,950,298 & 20 \\ \hline
    \end{tabular}
    }
\end{table}
\smartparagraph{Baselines.}
Our evaluation includes models across different domains, including DeepLight~\citep{deeplight} (Recommendation), Wide ResNet-101-2~\citep{wideresnet} (Vision), and TransformerXL~\citep{transformerxl} (Language modeling) as shown in Table \ref{tab:benchmark}.
We compare Global-QSGD against the no quantization baseline, QSGD (baseline for quantization), and PowerSGD (baseline for Allreduce compatible compression).
The evaluation is conducted with 3 random seeds, and the reported results are averages.

\smartparagraph{Convergence.}
   Figure \ref{fig:Loss} shows that Global-QSGD can achieve the fastest run time while preserving convergence with P2P. Both linear and exponential dithering achieve good convergence, while striking a trade-off between runtime and convergence performance. Specifically, as expected, linear dithering achieves the quickest run time; however, exponential dithering is slightly slower but achieves a better convergence similar to no quantization. In particular, exponential dithering can preserve better convergence than other compressors.  
   Given that NVLink P2P provides the highest bandwidth, Global-QSGD demonstrates even more significant speedup with lower-bandwidth network fabrics like SHM or inter-node settings.

\smartparagraph{Generalization.}
 We evaluate models with domain-specific metrics after the fixed-iteration training. Table \ref{tab:metric} shows that Global-QSGD can generally preserve generalizability as with no quantization.

\smartparagraph{Scalability.}
We also run a large-scale experiment with DeepLight on Google Cloud Platform (GCP), with 64 servers, each having one A100 GPU. The servers are connected through a shared network with bandwidth fluctuating from 200 Mbps to 1.5 Gbps.
We compare with the PowerSGD baseline as it is Allreduce compatible, while Allgather-based approaches, such as QSGD, are considerably slower and expensive, and thus not included.
Figure \ref{fig:DeepLightCloud} shows the training curve, where the model experiences more communication bottlenecks, allowing compression to achieve greater time savings.

\smartparagraph{Speedup.}
Figure~\ref{fig:speedup} shows the speedup obtained by Global-QSGD compared to the no quantization baseline when training different models for a constant number of epochs. The compression method usually obtains a higher speedup with slower network fabrics. Global-QSGD achieves up to 3.17$\times$ speedup in Cloud (GCP) and 3.51$\times$ with SHM. Even when communicating with P2P through NVLink, which has the highest bandwidth, Global-QSGD still achieves 1.38$\times$ speedup.
When applying exponential dithering on Wide ResNet-101-2 with P2P, a model which is not communication-bottlenecked with the highest bandwidth, Global-QSGD still delivers comparable performance with less than 2\% overhead.

\smartparagraph{Low bit-width.}
Table \ref{tab:4bitSpeedup} illustrates the speedup of $\GL^{q,p}_s$ relative to no quantization, highlighting Global-QSGD's adaptability to low bit-width quantization.
Our 4-bit implementation involves the overhead of packing 4-bit values into 8-bit. With dedicated hardware support, these performance gains could be further enhanced.

\smartparagraph{Limitations.}
The current implementation is not completely optimized. We make use of the NCCL P2P API, and there are synchronous invocations that prevent pipelining opportunities for further overlapping compression and communication.

\begin{table}[t]
    \centering
    \caption{Model quality after fixed-iteration training.}
    \vspace{6pt}
    \resizebox{0.45\textwidth}{!}{%
    \begin{tabular}{|l|c|c|c|}
    \hline
        ~ & \makecell{\textbf{DeepLight} \\AUC $(10^{-1})$} & \makecell{\textbf{Wide ResNet-101-2} \\Top 5} & \makecell{\textbf{TransformerXL}\\PPL}\\\hline
        No Quantization & $6.75\pm0.10$  & $87.06\%\pm0.04\%$ & $22.99\pm0.11$\\ \hline
        $\GL^{q,p}_s$ 8-bit & $6.95\pm0.03$ & $88.87\%\pm0.25\%$ & $32.35\pm0.71$\\ \hline
        $\GL^{q,p}_s$ 4-bit & $6.87\pm0.11$ & $88.57\%\pm0.30\%$ & $44.95\pm0.93$\\ \hline
        $\GE^{q,p}_s$ 8-bit & $6.87\pm0.03$ & $85.82\%\pm0.25\%$ & $23.68\pm0.12$\\ \hline
        QSGD 8-bit & $6.78\pm0.05$ & $89.39\%\pm0.32\%$& $30.82\pm0.38$ \\ \hline
        PowerSGD 32-rank & $6.76\pm0.06$& $88.19\%\pm0.22\%$ & $23.36\pm0.20$ \\ \hline
    \end{tabular}
    }
    \label{tab:metric}
    \vspace{1.5em}
\end{table}
\begin{table}[t]
    \centering
    \caption{Training throughput speedup normalized to no quantization.}
    \vspace{6pt}
    \resizebox{0.45\textwidth}{!}{%
    \begin{tabular}{|l|c|c|c|}
    \hline
         & \textbf{DeepLight} & \textbf{Wide ResNet-101-2} & \textbf{Transformer-XL} \\ \hline
        \multicolumn{4}{|c|}{\textbf{P2P Interconnect (NVLink)}} \\ \hline
        $\GL^{q,p}_s$ 8-bit & 1.38$\times$ & 1.21$\times$ & 1.05$\times$ \\ \hline
        $\GL^{q,p}_s$ 4-bit & 1.65$\times$ & 1.05$\times$ & 1.02$\times$ \\ \hline
        \multicolumn{4}{|c|}{\textbf{SHM Interconnect (PCIe)}} \\ \hline
        $\GL^{q,p}_s$ 8-bit & 3.51$\times$ & 2.72$\times$ & 1.73$\times$ \\ \hline
        $\GL^{q,p}_s$ 4-bit & 4.94$\times$ & 3.68$\times$ & 1.89$\times$ \\ \hline
    \end{tabular}
    }
    \label{tab:4bitSpeedup} 
\end{table}

\section{Conclusion}
We introduced Global-QSGD, a family of quantization operators for distributed learning based on global normalization. This enables the reduction of quantized values without changes of numeric representation (and corresponding overheads) with Allreduce communication and is backed by sound theory. Through extensive experiments, we showed that Global-QSGD yields fast convergence in realistic benchmarks. In particular, we demonstrated that Global-QSGD with exponential dithering achieves the best balance between the variance caused by compression and the speed of convergence.

\section*{Acknowledgments}
This publication is based upon work supported by the King Abdullah University of Science and Technology Research Funding (KRF) under Award No ORA-CRG2021-4699.
\bibliography{mybibfile}

\clearpage
\appendix

\onecolumn

\section{Proofs}
\label{app:proofs}
In this section, we include complete proofs of the claims made in the main paper.
\subsection{Proof of Lemma~\ref{lem:c_subset_d}}
Firstly, we show unbiasedness. 
\begin{align*}
    \E{\cQ(\xx)} &= \E{\frac{1}{n}\sum_{i=1}^n\cC_i(x_i)} = \frac{1}{n}\sum_{i=1}^n\E{\cC_i(x_i)}
    \overset{\eqref{eq:omega_quant}}{=} \frac{1}{n}\sum_{i=1}^n x_i = \bx.
\end{align*}
For the variance, 
\begin{align*}
    \E{\norm{\cQ(\xx) - \bx}_2^2} = \E{\norm*{\frac{1}{n}\sum_{i=1}^n\cC_i(x_i) - x_i}_2^2}
    = \frac{1}{n^2}\sum_{i=1}^n\E{\norm*{\cC_i(x_i)-x_i}_2^2} \overset{\eqref{eq:omega_quant}}{\leq} \frac{\omega}{n}\frac{1}{n}\sum_{i=1}^n \norm{x_i}_2^2,
\end{align*}
where the second equality is due to independence and zero mean of each summand.

\subsection{Proof of Lemma~\ref{lem:global_quant_reduction}}

It is easy to that such operator is unbiased since $\E{\rbr*{\xi(y_i)} } = \rbr*{\xi(y_i)}_j $ by construction. Therefore,
\begin{align*}
\E{\GQ^{q,p}_s(\xx)}& \overset{\eqref{eq:glob_quant}}{=} \E{\norm{\xx}_{q,p}  \frac{1}{n}\sum_{i=1}^n \signum(x_i) \circ \xi_i(y_i)}\\
&= \norm{\xx}_{q,p}   \frac{1}{n}\sum_{i=1}^n \signum(x_i) \circ \E{\xi_i(y_i)} \\
&\overset{\eqref{eq:rounding}}{=} \norm{\xx}_{q,p}   \frac{1}{n}\sum_{i=1}^n \signum(x_i) \circ y_i\\
&= \norm{\xx}_{q,p}  \frac{1}{n}\sum_{i=1}^n \signum(x_i) \circ \frac{x_i}{\norm{\xx}_{q,p}} = \bx.
\end{align*}

Furthermore, note that for $\GQ$, the local compressors do not belong to $\U^n(\omega)$ for any $\omega > 0$ due to their dependence on $\xx$. 

To obtain the variance bound, we show that it is sufficient to look at $n=1$, which corresponds to the unbiased compressor (Definition~\ref{def:omegaquant}), due to the following property.
\begin{align*}
    &\E{\norm*{\GQ^{q,p}_s(\xx) - \bx}_2^2} \displaybreak[1] \\ 
    &\overset{\eqref{eq:glob_quant}}{=} \E{\norm*{\norm{\xx}_{q,p} \frac{1}{n}\sum_{i=1}^n \rbr*{\signum(x_i) \circ \xi_i(y_i) - \frac{x_i}{\norm{\xx}_{q,p} }}}_2^2} \displaybreak[1] \\
    &= \frac{1}{n^2}\sum_{i=1}^n  \E{\norm*{\norm{\xx}_{q,p} \rbr*{\signum(x_i) \circ \xi_i(y_i) - \frac{x_i}{\norm{\xx}_{q,p} }}}_2^2} \displaybreak[1] \\
    &= \frac{1}{n^2}\E{\norm*{\norm{\xx}_{q,p} \signum(\xx) \circ \xi(\yy) - \xx}_2^2} \displaybreak[1] \\
    &= \frac{1}{n^2}\E{\norm*{\cQ^{q,p}_s(\xx) - \xx}_2^2}, 
\end{align*}
where the second inequality is due to independence of $\cbr*{\xi_i}$. If we can show that $\cQ^{q,p}_s(\xx) \in \U^{1,nd}(\omega)$, then 
\begin{align*}
    \E{\norm*{\GQ^{q,p}_s(\xx) - \bx}_2^2} &= \frac{1}{n^2}\E{\norm*{\cQ^{q,p}_s(\xx) - \xx}_2^2} \\
    \leq \frac{\omega}{n^2} \norm{\xx}_{2,2}^2 &= \frac{\omega}{n} \frac{1}{n}\sum_{i=1}^n \norm{x_i}_2^2,
\end{align*}
which implies $\theta = \nicefrac{\omega}{n}$.

\subsection{Proof of Theorem~\ref{thm:sparsity}}

The results provided in the theorem follow the same logic as we use in Lemma~\ref{lem:global_quant_reduction}, i.e., our compression is equivalent to applying standard compression to concatenated vector $\xx= \xx \eqdef \sbr*{x_1, x_2, \hdots, x_n}  \in \R^{nd}.$  

The rest of the proof uses known one-node results of \citet{qsgd2017neurips, horvath2019natural}. 

For the variance part, we firstly use the second part of Lemma~\ref{lem:global_quant_reduction}, which shows that $\cQ^{q,p}_s \in \cC(\omega) \Rightarrow \GQ^{q,p}_s \in \U^{n, d}(\theta)$ with $\theta = \nicefrac{\omega}{n}$. Secondly, we apply $n=1$ results of \citep[Theorem 7]{horvath2019natural} and \citep[Theorem 3.4]{qsgd2017neurips} combined with the fact about norm stating that $\norm*{\xx}_{q,p} \leq \norm{\xx}_{2,2}$ for $p,q \geq 2$. 

The claim about sparsity with $p=q=2$ follows from the first part of Lemma~\ref{lem:global_quant_reduction}, which implies that the number of non-zero elements of $\GQ^{q,p}_s$ before aggregation is the same as $\cQ^{q,p}_s$. The rest of the proof for $\cL^{2,2}_s$ follows directly from \citep[Lemma 3.1]{qsgd2017neurips}.  For $\cE^{2,2}_s$, one can also directly apply \citep[Lemma 3.1]{qsgd2017neurips} using the fact that the length of the first segment $[l_0, l_{1}]$ is $\nicefrac{1}{2^{s-1}}$.

\subsection{Proof of Lemma~\ref{lem:varianceBound}}
\label{app:variance}
As discussed in the main part of the paper, the only difference from the standard analysis is the difference in the variance bound. The lemma~\ref{lem:varianceBound} shows that both bounds are almost equivalent, where they only differ by a constant. Recall $\E{\cdot}$ denotes the standard expectation.
\begin{proof}
We assume that the noise due to sampling and compression are independent. For the first case, we have
\begin{align*}
    &\E{\norm*{\frac{1}{n}\sum_{i=1} \cC_i(\nabla f_i(x, \xi_i)) - \nabla f(x)}_2^2} \\
    &\quad =  \frac{1}{n^2} \sum_{i=1}  \E{\norm*{\cC_i(\nabla f_i(x, \xi_i)) - \nabla f_i(x)}_2^2} +  \frac{1}{n^2} \sum_{i\neq j}  \E{\dotprod{\cC_i(\nabla f_i(x, \xi_i)) - \nabla f_i(x)}{\cC_j(\nabla f_j(x, \xi_i)) - \nabla f_j(x)}} \\
    &\quad = \frac{1}{n^2} \sum_{i=1}  \E{\norm*{\cC_i(\nabla f_i(x, \xi_i)) - \nabla f_i(x)}_2^2} +  \frac{1}{n^2} \sum_{i\neq j}  \dotprod{\E{\cC_i(\nabla f_i(x, \xi_i)) - \nabla f_i(x)}}{\E{\cC_j(\nabla f_j(x, \xi_i)) - \nabla f_j(x)}}. \\
\end{align*}
Using tower property, we get 
\begin{align*}
    \E{\cC_i(\nabla f_i(x, \xi_i)) - \nabla f_i(x)} &= \EE{\xi_i \sim \cD_i}{\EE{\cC_i}{\cC_i(\nabla f_i(x, \xi_i)) - \nabla f_i(x) | \xi_i}} \\
    &= \EE{\xi_i \sim \cD_i}{\nabla f_i(x, \xi_i) - \nabla f_i(x)} = 0.
\end{align*}
Furthermore, using the same technique and the unbiasedness and bounded variance of the compression operator yields
\begin{align*}
    &\E{\norm*{\cC_i(\nabla f_i(x, \xi_i)) - \nabla f_i(x)}_2^2} = \EE{\xi_i \sim \cD_i}{\EE{\cC_i}{\norm*{\cC_i(\nabla f_i(x, \xi_i)) - \nabla f_i(x)}_2^2 | \xi_i}} \\
    &\quad = \EE{\xi_i \sim \cD_i}{\EE{\cC_i}{\norm*{\cC_i(\nabla f_i(x, \xi_i)) }_2^2|\xi_i}} - 2\EE{\xi_i \sim \cD_i}{\EE{\cC_i}{\dotprod{\cC_i(\nabla f_i(x, \xi_i))}{\nabla f_i(x)}| \xi_i}}  + \norm*{ \nabla f_i(x)}_2^2 \\
    &\quad = \EE{\xi_i \sim \cD_i}{\EE{\cC_i}{\norm*{\cC_i(\nabla f_i(x, \xi_i)) }_2^2|\xi_i}} - \norm*{\nabla f_i(x)}_2^2 \\
    &\quad = \EE{\xi_i \sim \cD_i}{\EE{\cC_i}{\norm*{\cC_i(\nabla f_i(x, \xi_i))- \nabla f_i(x, \xi_i) + \nabla f_i(x, \xi_i)}_2^2|\xi_i}} - \norm*{\nabla f_i(x)}_2^2 \\
    &\quad = \EE{\xi_i \sim \cD_i}{\EE{\cC_i}{\norm*{\cC_i(\nabla f_i(x, \xi_i))- \nabla f_i(x, \xi_i)}_2^2|\xi_i}} + \EE{\xi_i \sim \cD_i}{\norm*{ \nabla f_i(x, \xi_i)}_2^2}- \norm*{\nabla f_i(x)}_2^2 \\
    &\qquad + 2\EE{\xi_i \sim \cD_i}{\EE{\cC_i}{\dotprod{\cC_i(\nabla f_i(x, \xi_i)) - \nabla f_i(x, \xi_i)}{\nabla f_i(x, \xi_i)}| \xi_i}} \\
    &\quad = \EE{\xi_i \sim \cD_i}{\EE{\cC_i}{\norm*{\cC_i(\nabla f_i(x, \xi_i))- \nabla f_i(x, \xi_i)}_2^2|\xi_i}} + \EE{\xi_i \sim \cD_i}{\norm*{ \nabla f_i(x, \xi_i)}_2^2}- \norm*{\nabla f_i(x)}_2^2 \\
    &\qquad + 2\EE{\xi_i \sim \cD_i}{\dotprod{\EE{\cC_i}{\cC_i(\nabla f_i(x, \xi_i))} - \nabla f_i(x, \xi_i)}{\nabla f_i(x, \xi_i)}| \xi_i} \\
    &\quad = \EE{\xi_i \sim \cD_i}{\EE{\cC_i}{\norm*{\cC_i(\nabla f_i(x, \xi_i))- \nabla f_i(x, \xi_i)}_2^2|\xi_i}} + \EE{\xi_i \sim \cD_i}{\norm*{ \nabla f_i(x, \xi_i)}_2^2}- \norm*{\nabla f_i(x)}_2^2 \\
    &\qquad + 2\EE{\xi_i \sim \cD_i}{\dotprod{0}{\nabla f_i(x, \xi_i)}| \xi_i} \\
    &\quad \leq (\omega + 1) \EE{\xi_i \sim \cD_i}{\norm*{\nabla f_i(x, \xi_i)}_2^2} - \norm*{\nabla f_i(x)}_2^2 \\
    &\quad = (\omega + 1) \EE{\xi_i \sim \cD_i}{\norm*{\nabla f_i(x, \xi_i) - \nabla f_i(x) + \nabla f_i(x)}_2^2} - \norm*{\nabla f_i(x)}_2^2 \\
    &\quad = (\omega + 1) \EE{\xi_i \sim \cD_i}{\norm*{\nabla f_i(x, \xi_i) - \nabla f_i(x)}_2^2} + 2(\omega + 1) \dotprod{\EE{\xi_i \sim \cD_i}{\nabla f_i(x, \xi_i)} - \nabla f_i(x)}{\nabla f_i(x)}  + \omega \norm*{\nabla f_i(x)}_2^2 \\
    &\quad = (\omega + 1) \EE{\xi_i \sim \cD_i}{\norm*{\nabla f_i(x, \xi_i) - \nabla f_i(x)}_2^2} + \omega \norm*{\nabla f_i(x)}_2^2, 
\end{align*}
which concludes the first part of the proof. 
For the second part, we proceed analogously and have  that for $\cQ \in \U^{n, d}(\theta)$
\begin{align*}
    &\E{\norm*{\cQ\rbr*{\sbr{\nabla f_1(x, \xi_1), \ldots, \nabla f_n(x, \xi_n)}} - \nabla f(x)}_2^2}  \\
    &\quad = \EE{\cbr*{\xi_i \sim \cD_i}_{i=1}^n}{\EE{\cQ}{\norm*{\cQ\rbr*{\sbr{\nabla f_1(x, \xi_1), \ldots, \nabla f_n(x, \xi_n)}} - \frac{1}{n}\sum_{i=1}^n \nabla f_i(x, \xi_i)  + \frac{1}{n}\sum_{i=1}^n \nabla f_i(x, \xi_i) - \nabla f(x)}_2^2|\cbr*{\xi_i }_{i=1}^n}}\\
    &\quad = \EE{\cbr*{\xi_i \sim \cD_i}_{i=1}^n}{\EE{\cQ}{\norm*{\cQ\rbr*{\sbr{\nabla f_1(x, \xi_1), \ldots, \nabla f_n(x, \xi_n)}} - \frac{1}{n}\sum_{i=1}^n \nabla f_i(x, \xi_i)}_2^2|\cbr*{\xi_i }_{i=1}^n}} \\
    &\qquad + \EE{\cbr*{\xi_i \sim \cD_i}_{i=1}^n}{\norm*{\frac{1}{n}\sum_{i=1}^n \nabla f_i(x, \xi_i) - \nabla f(x)}_2^2}\\
    &\qquad +  \EE{\cbr*{\xi_i \sim \cD_i}_{i=1}^n}{\dotprod{\EE{\cQ}{\cQ\rbr*{\sbr{\nabla f_1(x, \xi_1), \ldots, \nabla f_n(x, \xi_n)}}|\cbr*{\xi_i }_{i=1}^n} - \frac{1}{n}\sum_{i=1}^n \nabla f_i(x, \xi_i)}{\frac{1}{n}\sum_{i=1}^n \nabla f_i(x, \xi_i) - \nabla f(x)}}.
\end{align*}
The first element of the scalar product is zero since $\cQ$ is unbiased. Therefore
\begin{align*}
    &\E{\norm*{\cQ\rbr*{\sbr{\nabla f_1(x, \xi_1), \ldots, \nabla f_n(x, \xi_n)}} - \nabla f(x)}_2^2}  \\
    &\quad = \EE{\cbr*{\xi_i \sim \cD_i}_{i=1}^n}{\EE{\cQ}{\norm*{\cQ\rbr*{\sbr{\nabla f_1(x, \xi_1), \ldots, \nabla f_n(x, \xi_n)}} - \frac{1}{n}\sum_{i=1}^n \nabla f_i(x, \xi_i)}_2^2|\cbr*{\xi_i }_{i=1}^n}} \\
    &\qquad + \EE{\cbr*{\xi_i \sim \cD_i}_{i=1}^n}{\norm*{\frac{1}{n}\sum_{i=1}^n \nabla f_i(x, \xi_i) - \nabla f(x)}_2^2}\\
    &\quad \leq \frac{\theta}{n}\sum_{i=1}^n\EE{\xi_i \sim \cD_i}{\norm*{\nabla f_i(x, \xi_i)}_2^2} + \EE{\cbr*{\xi_i \sim \cD_i}_{i=1}^n}{\norm*{\frac{1}{n}\sum_{i=1}^n \nabla f_i(x, \xi_i) - \nabla f(x)}_2^2}.
\end{align*}
For $\EE{\cbr*{\xi_i \sim \cD_i}_{i=1}^n}{\norm*{\frac{1}{n}\sum_{i=1}^n \nabla f_i(x, \xi_i) - \nabla f(x)}_2^2}$, we have
\begin{align*}
    &\EE{\cbr*{\xi_i \sim \cD_i}_{i=1}^n}{\norm*{\frac{1}{n}\sum_{i=1}^n \nabla f_i(x, \xi_i) - \nabla f(x)}_2^2} \\
    &\quad = \frac{1}{n^2}\sum_{i=1}^n\EE{\xi_i \sim \cD_i}{\norm*{\nabla f_i(x, \xi_i) - \nabla f_i(x)}_2^2} + \frac{1}{n^2}\sum_{i \neq j}\dotprod{\EE{\xi_i \sim \cD_i}{\nabla f_i(x, \xi_i)} - \nabla f_i(x)}{\EE{\xi_j \sim \cD_j}{\nabla f_j(x, \xi_j)} - \nabla f_j(x)} \\
    &\quad = \frac{1}{n^2}\sum_{i=1}^n\EE{\xi_i \sim \cD_i}{\norm*{\nabla f_i(x, \xi_i) - \nabla f_i(x)}_2^2} 
\end{align*}
Finally, for the $\EE{\xi_i \sim \cD_i}{\norm*{\nabla f_i(x, \xi_i)}_2^2}$, we have
\begin{align*}
    &\EE{\xi_i \sim \cD_i}{\norm*{\nabla f_i(x, \xi_i)}_2^2} = \EE{\xi_i \sim \cD_i}{\norm*{\nabla f_i(x, \xi_i) - \nabla f_i(x) + \nabla f_i(x)}_2^2} \\
    &\quad = \EE{\xi_i \sim \cD_i}{\norm*{\nabla f_i(x, \xi_i) - \nabla f_i(x) + \nabla f_i(x)}_2^2}  + \norm*{\nabla f_i(x)}_2^2 + 2\dotprod{\EE{\xi_i \sim \cD_i}{\nabla f_i(x, \xi_i)} - \nabla f_i(x)}{\nabla f_i(x)} \\
    &\quad = \EE{\xi_i \sim \cD_i}{\norm*{\nabla f_i(x, \xi_i) - \nabla f_i(x) + \nabla f_i(x)}_2^2}  + \norm*{\nabla f_i(x)}_2^2.
\end{align*}
Putting them all together
\begin{align*}
    &\E{\norm*{\cQ\rbr*{\sbr{\nabla f_1(x, \xi_1), \ldots, \nabla f_n(x, \xi_n)}} - \nabla f(x)}_2^2}  \\
    &\quad \leq \frac{\theta}{n}\sum_{i=1}^n\EE{\xi_i \sim \cD_i}{\norm*{\nabla f_i(x, \xi_i)}_2^2} + \EE{\cbr*{\xi_i \sim \cD_i}_{i=1}^n}{\norm*{\frac{1}{n}\sum_{i=1}^n \nabla f_i(x, \xi_i) - \nabla f(x)}_2^2} \\
    &\quad = \frac{\theta}{n}\sum_{i=1} \E{\norm*{\nabla f_i(x))}_2^2}  + \rbr*{\theta + \frac{1}{n}} \frac{1}{n}\sum_{i=1}\E{\norm*{\nabla f_i(x, \xi_i) - \nabla f_i(x)}_2^2}
\end{align*}
yields the desired bound.
\end{proof}

For the adaptive methods, we show that the assumptions in Section 2.3 of \citep{defossez2022a} remain applicable when combining stochastic gradients with global quantization. The only assumption related to the stochastic gradients is that the $\ell_{\infty}$ norm of the stochastic gradients is almost surely bounded. 

In our case, the stochastic estimator has the following form
\begin{align*}
    \cQ\rbr*{\sbr{\nabla f_1(x, \xi_1), \ldots, \nabla f_n(x, \xi_n)}}.
\end{align*}

We assume that $\|\nabla f_i(x, \xi_i)\|_{\infty} \leq R$, for all $i \in [n].$ Furthermore, for both linear and exponential global quantization, we have 
\begin{align*}
    \|\cQ\rbr*{\sbr{\nabla f_1(x, \xi_1), \ldots, \nabla f_n(x, \xi_n)}}\|_{\infty} &= \left\|\frac{1}{n} \sum_{i=1}^n \rbr*{\cQ^{q,p}_s(\sbr{\nabla f_1(x, \xi_1), \ldots, \nabla f_n(x, \xi_n)})}_i\right\|_{\infty} \\
    &\leq \frac{1}{n} \sum_{i=1}^n \left\| \rbr*{\cQ^{q,p}_s(\sbr{\nabla f_1(x, \xi_1), \ldots, \nabla f_n(x, \xi_n)})}_i\right\|_{\infty} \\
    &\leq \|\sbr{\nabla f_1(x, \xi_1), \ldots, \nabla f_n(x, \xi_n)}\|_{p,q} \\
    &= \rbr*{\sum_{i=1}^n \norm{\nabla f_i(x, \xi_i)}_q^p}^{1/p} \\
    &\leq \rbr*{\sum_{i=1}^n (d\|\nabla f_i(x, \xi_i)\|_{\infty})^{p/q}}^{1/p} \leq n^{1/p} d^{1/q} R.
\end{align*}
Therefore, global quantization preserves the bound of the $\ell_{\infty}$ norm of the stochastic gradients up to a constant. This implies that all the convergence guarantees of \citet{defossez2022a} also apply to our setting with global quantization. 

\subsection{Proof for Equation~\ref{eq:condition}}
\label{app:modeling}
\begin{align*}
            2\log(N)\alpha + 2\frac{\log(N)S}{\beta} + \frac{\log(N)S}{\gamma} &> \displaybreak[1]2\log(N)\alpha + 2\frac{\log(N)\hat{S}}{\beta} + \frac{\log(N)\hat{S}}{\hat{\gamma}} + {\delta}S \displaybreak[1]\\
    2\frac{S}{\beta} + \frac{S}{\gamma} &> 2\frac{\hat{S}}{\beta} + \frac{\hat{S}}{\hat{\gamma}} \displaybreak[1]\\
    2\frac{\rho}{\beta} + \frac{\rho}{\gamma} &> 2\frac{1}{\beta} + \frac{1}{\hat{\gamma}} \displaybreak[1]\\
    2\rho\gamma\hat{\gamma} + \rho\beta\hat{\gamma} &> 2 \gamma\hat{\gamma} + \beta\gamma \displaybreak[1]\\
    \rho\beta\hat{\gamma} - \beta\gamma &> 2 \gamma\hat{\gamma} - 2\rho\gamma\hat{\gamma} \displaybreak[1]\\
    \beta(\rho\hat{\gamma}-\gamma) &> 2 \gamma\hat{\gamma} - 2\rho\gamma\hat{\gamma} \displaybreak[1]\\
\end{align*}
Since $\hat{\gamma}$ is the computation speed (byte/s), so $\hat{\gamma}\geq0$. We divide both sides of the inequality by $\hat{\gamma}$, and replace $\omega=\frac{\gamma}{\hat{\gamma}}$:
\begin{align*}
    \beta(\rho-\omega) &> 2 \gamma - 2\rho\gamma\\
    \beta(\rho-\omega) &> 2 \gamma (1-\rho)\\
\end{align*}
Then we have:
\begin{equation*}
\left\{ 
\begin{array}{ll}
\beta > \frac{2\gamma(1-\rho)}{(\rho-\omega)}, & \text{if } (\rho > \omega), \\
\beta < \frac{2\gamma(1-\rho)}{(\rho-\omega)}, & \text{if } (\rho < \omega).
\end{array} 
\right.
\end{equation*}

\clearpage

\section{Bandwidth Measurement}
\label{app:bandwidth}
We begin by describing our hardware setup. Our experiments were conducted on a node equipped with four A100-SXM4-80GB GPUs. According to NVIDIA's specifications, each A100 GPU can achieve up to 300 GB/s (600 GB/s bi-directional) bandwidth using 12 channels of 4th-generation NVLink.

To verify the bandwidth, we ran the following command:

\begin{lstlisting}[basicstyle=\ttfamily\small]
nvidia-smi nvlink --status
GPU 0: NVIDIA A100-SXM4-80GB (UUID: GPU-d9dba6f4-6927-ad43-418f-6d87f2265a79)
Link 0: 25 GB/s
Link 1: 25 GB/s
Link 2: 25 GB/s
Link 3: 25 GB/s
Link 4: 25 GB/s
Link 5: 25 GB/s
Link 6: 25 GB/s
Link 7: 25 GB/s
Link 8: 25 GB/s
Link 9: 25 GB/s
Link 10: 25 GB/s
Link 11: 25 GB/s
\end{lstlisting}

The output confirms that each GPU is connected to 12 NVLink channels, each providing 25 GB/s of single-direction bandwidth, resulting in a total of 300 GB/s.

Next, we used the following command to examine the peer-to-peer (P2P) connections between GPUs:

\begin{lstlisting}[basicstyle=\ttfamily\small]
nvidia-smi topo -m
GPU0 GPU1 GPU2 GPU3 NIC0 NIC1 CPU Affinity NUMA Affinity GPU NUMA ID
GPU0 X NV4 NV4 NV4 SYS SYS 0-127 0 N/A
GPU1 NV4 X NV4 NV4 SYS SYS 0-127 0 N/A
GPU2 NV4 NV4 X NV4 SYS SYS 0-127 0 N/A
GPU3 NV4 NV4 NV4 X PHB PHB 0-127 0 N/A
NIC0 SYS SYS SYS PHB X PIX
NIC1 SYS SYS SYS PHB PIX X
\end{lstlisting}

This confirms that the GPUs are connected via NVLink with 12 channels distributed among the three peers, resulting in each GPU pair being connected with four NVLink channels, accounting for a 100 GB/s single-direction bandwidth.

We measured the GPU-to-GPU bandwidth with the following command:
\begin{lstlisting}[basicstyle=\ttfamily\small]
mpirun -np 10 ./build/all_reduce_perf -b 25M -e 25M -g 4
\end{lstlisting}

The bandwidth is obtained using the NVIDIA NCCL Test suite, a standard tool for testing bandwidth with NVIDIA devices. We performed the NCCL Allreduce operation with a message size of 25 MB, consistent with our deep learning workload communication settings (the default bucket size in PyTorch DDP). It is important to note that Allreduce involves multi-stage collective communication, which does not always saturate the bandwidth and therefore does not achieve peak P2P throughput. Therefore, we empirically measure the bandwidth rather than using theoretical limits. 

We conducted measurements under two settings:
With Nvidia GPU Direct (P2P) enabled, where data transfer occurs over NVLink.
Using Shared Memory (SHM), where data is transferred through the host memory before reaching the destination.
We controlled these settings by modifying NCCL's environment variable NCCL\_P2P\_DISABLE. The measured results were 53.9 GB/s for P2P and 5.4 GB/s for SHM.

\end{document}